# Developing Brain Atlas through Deep Learning

Asim Iqbal[1,2], Romesa Khan[3], Theofanis Karayannis[1,2]

[1]Laboratory of Neural Circuit Assembly, Brain Research Institute (HiFo), UZH
[2]Neuroscience Center Zurich (ZNZ), UZH/ETH Zurich
[3]Department of Biology (D-BIOL), ETH Zurich

*Correspondence should be addressed to T.K. (karayannis@hifo.uzh.ch)

## Abstract

Neuroscientists have devoted significant effort into the creation of standard brain reference atlases for high-throughput registration of anatomical regions of interest. However, variability in brain size and form across individuals poses a significant challenge for such reference atlases. To overcome these limitations, we introduce a fully automated deep neural network-based method (*SeBRe*) for registration through **Se**gmenting **B**rain **Re**gions of interest with minimal human supervision. We demonstrate the validity of our method on brain images from different mouse developmental time points, across a range of neuronal markers and imaging modalities. We further assess the performance of our method on images from MR-scanned human brains. Our registration method can accelerate brain-wide exploration of region-specific changes in brain development and, by simply segmenting brain regions of interest for high-throughput brain-wide analysis, provides an alternative to existing complex brain registration techniques.

# Introduction

With the development and efficient implementation of various methods in neuroscience for labelling specific populations of brain cells *in situ*, such as the generation of genetically modified animals, or the more conventional immunocytochemistry and mRNA in-situ hybridization (ISH) on brain tissue, neuroscientists are able to label and track various neuronal types in different brain regions across development [1]. These cellular expression patterns are captured through a variety of high-throughput imaging techniques, such as whole tissue light-sheet microscopy, or bright wide-field and fluorescent microscopy using high-resolution confocal microscopes and slide scanners, which allow for their exploration at a meso-scopic level. However, quantitative analysis of these brain datasets remains a great challenge in the field of neuroscience, with a major issue being the complexity of registering mouse brain sections against a standard reference atlas.

A number of efforts are underway to develop high-throughput image registration frameworks for analysing such large-scale brain datasets [2, 3]. Nevertheless, most of these frameworks are semi-manual, such that the user is either expected to set a certain range of parameters like intensity threshold, background contrast, etc. for every brain section or even completely transform their brain datasets into the framework-readable format. Notwithstanding that, these methods result in limited performance in registration of every brain section against a reference atlas, hence lacking the generalizability for analysing a variety of datasets. In addition and importantly, there also are no reference atlases for most of the developing age groups.

The development of deep learning (DL)-based techniques is now providing state-of-the-art results in real-world object classification [4], localization [5] and segmentation [6] tasks. Even though DL methods have been applied to different body organs such as the heart and bones for segmentation [7], their usage in the analysis of complex brain image datasets has been scarce [8, 9] and not applied to atlas registration and region annotation (classification). Amongst the many challenges, this Artificial Intelligence (AI)-based approach would require the generation of a large enough dataset of brain-sections that is *1)* labelled with brain regions in reference to a standard atlas and *2)* captures the variability of various regions in different sections across the brain.

We propose that the task of registering images of brain sections against a standard reference atlas can also be achieved by segmenting the regions in the brain through feature engineering, using a deep neural network (DNN). Therefore, we introduce a different approach to the classical problem of brain image registration, which relies on '*registration through segmentation*'. This approach deploys a fully automated DNN-based method to segment and annotate various regions in images of the mouse brain at different ages, as well as in images of the human brain, by optimization of the Mask R-CNN architecture [6] and applying transfer learning on a network pre-trained on the MS COCO dataset [10]. To train and test the performance of the network, we generated two distinct human-annotated mouse brain region datasets using the Allen Brain Institute Online Public Resource (OPR) [11], as well as a set of publicly available human MR images [14]. By comparing the performance of our network with human-annotated ground-truth data and traditional brain registration and segmentation approaches, we achieve high average precision (AP) scores and demonstrate the power of our method in comparison to the traditional brain registration and segmentation approaches. In essence, *SeBRe* automatically *generates* the brain reference atlas corresponding to any input brain section. This approach marks a paradigm shift in dealing with large scale brain datasets, since it is independent of *1)* the time-intensive step of manual selection, followed by *2)* the computationally expensive step of registration, of the putative reference atlas onto the input brain sections. We propose that our approach of segmenting brain regions can become a launch pad for replacing the classical methods for registering brain sections against a standard reference atlas in the future.

## Results

**Performance of *SeBRe* on mouse brain images.** The block diagram summary of our approach is presented in **Fig. 1**, where an input brain image from a developing mouse brain (left) are fed into the DNN and after passing through a series of feature processing stages, brain regions are localized, classified and segmented (right). Step-by-step processing of a sample input brain section is demonstrated in **Fig. 2**, where an input brain section is passed through a series of image processing stages where brain regions are proposed, classified and segmented. We demonstrate the ground-truth dataset generation in **Supplementary Figs. 1-3**, explained in the Methods section. To facilitate the neuroscience community, we open-source our pre-trained deep neural network on the mouse and human datasets, along with the complete dataset.

We tested the performance of *SeBRe* by designing two kinds of experiments: *1)* training and testing the DNN on images of two ISH brains of an intermediate mouse age (P14) to check the scalability of our method in segmenting brain regions of earlier (P4) and later (P28, P56) post-natal time points, and *2)* training and testing the DNN on an extended mouse dataset of nine ISH brains - covering three post-natal mouse ages (P4, P14, P56), from different genetically modified animals. We use the trained model from (2) for demonstrating the generalizability of our method to segmenting regions in developing mouse brain images, captured through different imaging modalities, such as FISH (fluorescent in situ hybridization).

*SeBRe*'s output on various sections of P14 mouse brains is demonstrated in **Fig. 3**, where the first column shows the human expert-annotated masks for eight regions of interest, in randomly selected mouse brain sections labelled with Glutamate Decarboxylase 1 (GAD1) and Vesicular GABA Transporter (VGAT) neuronal markers. The middle column shows the performance of *SeBRe* in segmenting the eight regions in these brain sections. The right-most column shows the performance of the network on rotated [−20°, 20°] versions of the same brain images. The network performs well in segmenting brain regions in upright as well as rotated versions of the brain images. We further test the performance of *SeBRe* on rotated brain sections (**Fig. 4**) drawn from the extended mouse dataset to demonstrate that the method is invariant to the rotation of mouse brain images. This approach is useful for registering brain sections in real-time under the microscope where the brain tissues could be placed at various angles of convenience.

The mean AP scores achieved for brain sections in the original (P14) and the extended (P4, P14, P56) test datasets are 0.84 and 0.87, respectively. Performance of the network is optimal in most brain regions, including isocortex, hippocampus, basal ganglia, telencephalic vesicle and the midbrain. For segmenting the pre-thalamus, the performance of the network is limited due to a large variation in the structure and size of this region, as we move from lateral to medial images, across the sagittal plane.

**Performance of *SeBRe* on mouse brain images labelled with previously "unseen" markers and ages.** After training and testing the performance of our network on P14 brain sections, we further checked the scalability of our method by testing it on various mouse brain images from different developmental time points (P4, P14, P28 and P56), for commonly used neuronal markers that the network had not been trained on (CaMKIIa, Nissl, GAD1 and VGAT) [11]. $Ca^{2+}$/calmodulin-dependent protein kinase IIa (CaMKIIa) labels the largest population of neurons in the brain called excitatory (Glutamatergic) neurons, GAD1 and VGAT label the second largest neural population called inhibitory (GABAergic) neurons, and Nissl labels all the cells in the brain and is often used to explore the morphology of brain tissue. **Supplementary Fig. 4** shows the results of segmentation on various mouse brain section images, across different ages and neuronal markers. Performance of *SeBRe* on randomly selected brain sections from a P4 mouse brain of GAD1 and VGAT is shown in **Supplementary Fig. 4 (a-b)**. The network seems to perform optimally in segmenting isocortex, thalamus and the telencephalic vesicle, whereas detection of the midbrain and hindbrain regions proves a challenge for the network. **Supplementary Fig. 4 (c-d)** shows the performance of the *SeBRe* on randomly selected brain sections from Nissl and CaMKIIa brains at P4. Although, the network has been trained on GAD1 and VGAT tissue sections of only a single age (P14), it performs equally well in segmenting brain regions, for various neuronal markers at different developmental ages. The segmentation results for P28+ mouse brain regions in GAD1, VGAT, CaMKIIa and Nissl brains are shown in **Supplementary Fig. 4 (g-j)**. Similar to the P14 testing dataset, in the medial brain sections of older mice, the network seems to occasionally omit a few regions, such as the midbrain and pre-thalamus, but the segmentation of isocortex and other regions remains good.

**Performance of *SeBRe* on mouse brain images acquired through a previously "unseen" imaging modality.** We tested the performance of SeBRe, trained on the extended mouse dataset, on a variety of fluorescently-labelled brain images from various transgenic mouse lines which label different brain areas and cell types, hence covering a diverse range of image intensity profiles [11]. These mouse lines are genetically modified using a Cre-LoxP system, with a variety of Cre driver lines recombining a red reporter allele and hence labelling different cell populations (represented in *Probe/Driver-Cre-Reporter* notation). In addition, many of them also display a second color (green) after an in-situ hybridization against GAD1, Slc17a6 or Rorb. We tested the performance of *SeBRe* in segmenting brain regions of the mouse brain images from the following transgenic mouse brains: GAD1/Cux2-CreERT2-Ai14(tdTomato), GAD1/Grik4-Cre-Ai14(tdTomato), Rorb/Scnn1a-Cre-Ai14(tdTomato), Slc17a6/Slc32a1-Cre-Ai14(tdTomato) and GAD1/Gpr26-CreKO250-Ai14(tdTomato). Although only trained on GAD1 or VGAT ISH images, *SeBRe* performs reasonably well in segmenting mouse brain regions in FISH brains, imaged under a fluorescent microscope (**Fig. 5**), even in cases when the brain region boundaries are not clearly defined (**Fig. 5 (f-g)**) or broken (**Fig. 5 (a, k)**). It is interesting to observe that *SeBRe* performs well, independent of the brain imaging modality, as the DNN is able to detect the features of interest without requiring any additional training on a completely separate image set it was never trained on. This stresses upon the power of an AI-based segmentation method over traditional registration techniques.

**Performance of *SeBRe* in fine-scale segmentation of mouse hippocampus.** To evaluate if our method can also be extended to handle the more complex task of fine-scale sub-region segmentation within a brain tissue of interest, we tested the performance of *SeBRe* in isolating the four major sub-regions of the mouse hippocampus, CA1, CA2, CA3 and the dentate gyrus (DG), in a Nissl-labelled ISH brain from the Allen Brain OPR. We trained the network on 3/4 dataset, selected randomly, and subsequently tested the performance on the remaining 1/4. The results are demonstrated in **Fig. 6** on three sample coronal mouse brain sections. We further test the performance of *SeBRe* on rotated samples of coronal brain sections, with results demonstrated in **Supplementary Fig. 5**, for different orientations of two randomly selected rostral (**Supplementary Fig. 5 (a)**) and caudal (**Supplementary Fig. 5 (b)**) brain sections, respectively. Sub-region masks predicted by the network adequately match the human-annotated ground-truth segmentations for all four sub-region classes, in rostral as well as caudal brain sections containing the hippocampus structure. The network has a high mean AP score of 0.9 in detecting and distinguishing sub-regions of the mouse hippocampus. The network is also able to identify and segment the major sub-regions of the hippocampus in ISH brain sections labelled with a range of other neuronal markers spanning diverse spatial expression patterns e.g. performance of *SeBRe* on a coronal section labelled with Neurogranin (Nrgn) marker is shown in **Supplementary Fig. 6 (b)**. We further tested if our method is also generalizable to brain images captured under a previously 'unseen' imaging modality. We observe reasonable performance of *SeBRe*, which was trained on ISH brain images, on a FISH transgenic mouse brain, simultaneously expressing a pan-neuronal red reporter (NeuN) and a green fluorescent marker for cell bodies and neurites (NF-160), as illustrated in **Supplementary Fig. 6 (a)**, despite significant differences from the bright-field image intensity range, on which the network is originally trained.

**Performance of *SeBRe* on human brain images.** In order to evaluate the generalizability of our deep-learning based registration method to human data as well as other imaging modalities, we further tested the performance of *SeBRe* on an open-source, manually-segmented human T1-weighted magnetic resonance imaging (MRI) brain dataset from the Internet Brain Segmentation Repository (IBSR) [14]. We trained the network on 2/3 of the dataset, randomly selected, and subsequently tested the performance on the remaining 1/3. Randomly selected horizontal sections, drawn at different dorso-ventral planes, from the scanned brain volumes, are shown in **Fig. 7 (a)**. The human-annotated ground-truth masks for eight different subcortical structures, overlaid onto the brain sections are shown in **Fig. 7 (b)**, whereas **Fig. 7 (c)** shows the segmentation performance of *SeBRe* on these input brain sections. The network output closely matches the human-annotated ground-truth segmentations for all eight subcortical regions, in both dorsal as well as ventral sections, giving a high mean AP score of 0.95.

**Comparison of *SeBRe* with image processing-based brain registration techniques.** To demonstrate the power of our '*registration through segmentation*' approach, the performance of *SeBRe* was compared with two other commonly used image registration methods: elastix, a toolbox for rigid and non-rigid registration of medical images [12], and the Neurodata Registration module (ndreg), which uses affine and non-affine transformations to align a mouse brain reference atlas image to the brain section images [13]. For a fair comparison, mouse brain reference atlas images, comprising only of the eight regions of interest on which *SeBRe* was trained were registered onto the corresponding mouse brain sections, using ndreg and elastix. **Fig. 8 (a)** visually demonstrates the registration performance of *SeBRe*, elastix and ndreg on lateral and medial sagittal sections from GAD1 and VGAT brains. Mean squared error (MSE) score for the 'registered' masks returned by *SeBRe* was notably lower than the MSE scores for both elastix and ndreg, as shown in **Fig. 8 (b-c)**, indicating higher registration accuracy achieved by our method. It is interesting to observe that these two commonly used registration techniques seem to underperform on the same type of images, as indicated by the increased MSE score in registering rotated samples, in contrast to our machine intelligence-based method, *SeBRe* (**Supplementary Fig. 7**).

**Comparison of *SeBRe* with other deep learning-based brain segmentation techniques.** To demonstrate the robustness of our segmentation approach, the performance of *SeBRe* was also compared with previously built deep learning-based segmentation methods, including DeepLab [15], BrainSegNet [16] and QuickNAT [17], across the eight subcortical regions of the IBSR dataset. The region-wise average DICE coefficient score is reported for all methods, in **Table 1 (column # 6-9)**, where it can be appreciated that *SeBRe* gives the highest segmentation accuracy among the evaluated methods in five out of eight classes (subcortical regions), and comparable accuracy for the remaining classes. The Hausdorff distance and contour mean distance (CMD) scores for each subcortical region are further compared between *SeBRe* and DeepLab, in **Table 1 (column # 2-5)**, where *SeBRe* provides consistently lower distance scores for segmentation of all the regions.

In conclusion, we introduce a deep neural network-based method called *SeBRe*, to classify and segment different regions in mouse and human brain images. To test the performance of our method, we utilize the open-source Allen Brain ISH mouse brain images, and generate a human-annotated dataset of nine brains from three postnatal ages: P4, P14 and P56. We demonstrate proper and accurate segmentation of eight regions in mouse brain images, from lateral and medial sagittal planes. Notably, our method is scale invariant i.e. it performs equally well in segmenting brain images that vary substantially in the relative sizes of the brain regions. This is crucial not only for accurately segmenting regions at different stages of development, but also in tissue sections in which the proper geometry of the brain has been compromised due to any methodological issues (e.g. during brain tissue slicing), or even pathological deformities (e.g. neurodegeneration in Alzheimer disease). This is demonstrated by the fact that although *SeBRe* was trained only on a P14 image dataset, the network performs equally well on other developmental ages including P4, P28 and P56. We further demonstrate the performance of our method on brain sections obtained through a different imaging modality such as FISH, pointing towards the generalizability of *SeBRe* in segmenting brain regions in a variety of bioimaging modalities.

Moreover, we demonstrate the performance of *SeBRe* in fine-segmentation of the brain sub-regions in mouse hippocampus across different neuronal markers as well as different imaging modalities i.e. ISH and FISH. In addition, we show that the performance of *SeBRe* is extendable to other species, by testing it on a publicly available human MR imaging dataset. We therefore believe that *SeBRe* can become a valuable tool for many bioimaging applications, such as segmenting brain regions in microscopy images for brain-wide analysis in animal models, or even human MRI and post-mortem tissue. Finally, although we demonstrate the application of our method in generating a reference atlas for mouse and human brain regions and sub-regions, it is noteworthy that, in essence, a similar approach could also be adapted for any other primate or non-primate model organism, such as monkey or drosophila.

**Author contributions**
A.I., R.K. and T.K. conceptualized the study and wrote the paper. A.I and R.K. developed the *SeBRe* method and performed the quantitative comparison with other registration and segmentation methods.

**Competing interests**
The authors declare no competing interests.

## Methods

**Generating the ground-truth dataset for mouse brain.** We generate the ground-truth dataset by first fetching the open-source In-Situ Hybridized (ISH) mouse brain images for two different genetic markers, GAD1 and VGAT, from the Allen Brain OPR [11]. These two markers capture the population of GABAergic neurons in developing mouse brains. We choose P14 as the intermediate age of animal that captures enough variance in terms of brain region development between a young animal pup (P4) and an adult mouse (P28+). These brain sections are cut at 20μm-thick in sagittal planes and each is 200 microns apart from the next, covering one hemisphere of a whole brain from lateral to medial. **Supplementary Fig. 1** illustrates the process of ground-truth data generation on a sample GAD1 mouse brain section. **Supplementary Fig. 2** shows the sample human-annotated sagittal sections of a complete GAD1 brain at P14. The brain sections are overlaid by the Allen developing mouse brain reference atlas, with each region assigned a unique color code. Human experts manually registered the brain sections with scalar vector graphics (SVG) files of the reference atlas, using Boxy SVG editor. The SVG files of developing mouse brain atlas were imported from Allen Brain OPR. GAD1 and VGAT brains at P14 consist of 36 (17 and 19 respectively) brain sections. Six sections were removed from these two brains, as these did not meet the required quality criteria due to broken/damaged tissue. To increase the variability of the sample images, for each brain section, we introduce a synthetic variance of a $2°$ rotation, in the range of $[-20, 20]$, resulting in 20 rotated versions per brain section. We apply the same procedure for all 30 brain sections with the resulting augmented dataset comprising of 600 brain sections. We apply a down-sampling ratio of 25% on each image in the dataset to reduce the computational cost. Out of these 30 sections, we randomly choose 2/3 of the brain sections with their rotated versions (400 images) for training and the remaining 1/3 with their rotated versions (200 images) for testing.

**Generating masks for brain regions of mouse.** To demonstrate the performance of *SeBRe*, we choose eight major regions in the developing mouse brain for training and testing, namely: isocortex, hippocampus, basal ganglia, thalamus, prethalamus, midbrain, telencephalic vesicle (olfactory bulb and partial forebrain) and hindbrain (cerebellum). Sample binary masks of five example regions, along with their corresponding brain sections, are shown in **Supplementary Fig. 3**. These masks are generated from the human expert-annotated ground-truth dataset, where each brain region is masked with a unique color code, as shown in **Supplementary Fig. 2**.

**Generating the extended ground-truth dataset for mouse brain across different ages.** To scale our brain registration approach to different mouse developmental ages, we generate an extended ground-truth dataset, comprised of ISH mouse brains from the Allen Brain OPR, labelled with different neuronal markers: GAD1, VGAT and Nissl, across three distinct developmental time-points, P4, P14 and P56 (total nine brains). Each human annotated brain section is overlaid with the manually registered Allen developing mouse brain reference atlas for the corresponding age. Ground-truth binary masks for the eight brain regions of interest are generated using unique colour codes for each region in the reference atlas, as explained above.

To increase training data variability, we apply image augmentation by introducing a synthetic variance of a $2°$ rotation with a range of $[-20, 20]$ for each brain section after down-sampling by a factor of 16 (6.25%). To demonstrate the generalizability of our registration framework across different ages, as well as different neuronal markers, we train the network on the complete Nissl and VGAT brains, of P4, P14 and P56 mice, and test the network on completely different brains, labelled for GAD1, across P4, P14 and P56 ages.

**Generating the ground-truth dataset for mouse hippocampus.** To demonstrate the performance of the network on an even finer scale, in segmenting sub-regions of a brain region, we generate a ground-truth dataset comprised of an ISH adult mouse brain from the Allen Brain OPR, labelled with Nissl. The coronal brain sections are evenly spaced at 100μm apart, covering the complete brain, from rostral to caudal. Only 27 coronal brain sections that contain the hippocampus structure are included in the dataset. Each coronal brain section is overlaid by the matching reference section from the Allen adult mouse brain coronal reference atlas, registered onto the right hemisphere of the Nissl brain. This atlas was mirrored and manually registered by the human experts also onto the left hemisphere. Ground-truth binary masks for the four major sub-regions of the mouse hippocampus, CA1, CA2, CA3 and DG, are generated using the unique color codes for each sub-region in the reference atlas, as explained above.

To enhance training data variability, we apply image augmentation through synthetic variation: each coronal brain section was horizontally flipped, followed by rotation with a step of 2° in the range [−20, 20]; creating 20 rotated versions of each brain section, as well as the flipped counterpart. Each brain section image in the dataset, is down-sampled by a factor of 16 (6.25 %). To ensure that training and testing datasets are non-overlapping, we train the network on a 800 brain sections and test on 280 brain sections, where both sets are mutually exclusive. **Generating the ground-truth dataset for human brain.** The ground-truth dataset is generated from the publicly-available 18 T1-weighted MRI brain scans provided by the IBSR [14]. The scanned brains belong from 7 to 17-year old human subjects, therefore capturing adequate variability across different ages. We use a 'cropped' version $(158 \times 123 \times 145)$ of the original IBSR 3D volumes as adapted in [15], creating a 2D dataset by drawing brain sections in the horizontal plane. Manually-segmented labels are provided for 32 regions in the original dataset, of which we choose a subset of 8 prominent subcortical structures: caudate, pallidum, putamen and thalamus, in both hemispheres. These 8 regions are selected to enable a direct comparison with other deep learning-based segmentation methods, including DeepLab. Ground-truth binary masks for each region of interest are generated from the human-annotated brain volume corresponding to each MR scanned brain, where each brain region is masked with a unique colour code. A subset of brain sections with empty masks were removed from the dataset. From the complete dataset of 557 brain sections, 2/3 of the brain sections (372 images) are used for training the network, and the remaining 1/3 (185 sections) are utilized for testing.

**Network architecture.** *SeBRe* is designed by optimization of Mask R-CNN architecture, constructed using a convolutional backbone that comprises of the first five stages of the very deep ResNet101 [18] and Feature Pyramid Network (FPN) [19] architectures. Our network architecture is shown in **Supplementary Fig. 8**. The feature map is processed by a Region Proposal Network (RPN), which applies a convolutional neural network over the feature map in a sliding-window fashion. The RPN segregates and forwards the predicted $n$ potential Regions of Interest (RoI) from each window to the Mask R-CNN 'heads' based on the FPN. The RoI feature maps undergo a critical feature pooling operation by a pyramidal RoIAlign layer, which preserves a pixel-wise correspondence to the original image. Each level of the pyramidal RoIAlign layer is assigned a RoI feature map from the different levels of the FPN backbone, depending on the feature map area, returning $n$ pooled feature maps, $P_n[7 \times 7]$. Three arms of the FPN perform the core operations of brain region segmentation. The 'classifier' and 'regressor' heads, inherited from the Faster R-CNN [5], detect and identify distinct brain regions, and compute region-specific bounding boxes. The classifier output layer returns a discrete probability distribution $[n, 9]$, for nine different object classes (eight brain regions + background). The regressor output layer gives the four (x-coordinate, y-coordinate, width, height) bounding-box regression offsets to be applied for each class, per RoI $[n, (4 \times 8)]$. **Fig. 2** illustrates the step by step procedure of brain region localization and classification, performed by the FPN heads 'classifier' and 'regressor' heads.

A fully convolutional network (FCN) forms the more recent mask-prediction arm, returning a binary mask spanning each segmented brain region. The mask arm applies a mask of resolution

$m \times m$ for each class, for each RoI $[n, (8 \times m^2)]$. Respectively, the output of the backbone architecture and the mask head, for a single brain section is shown in **Supplementary Fig. 9**. The network is trained using a stochastic gradient descent algorithm that minimizes a multi-task loss ($L$) corresponding to each labelled RoI:

$$L = L_{cls} + L_{reg} + L_{mask}$$

where $L_{cls}$, $L_{reg}$ and $L_{mask}$ are the region classification, bounding box regression and predicted masks' loss, respectively, as defined below.

$$L(p_i, q_i) = \frac{1}{n_{cls}} \sum_i L_{cls}(p_i, p_i^*) + \mu \frac{1}{n_{reg}} \sum_i L_{reg}(q_i, q_i^*)$$

$p_i$ is the probability of the $i^{th}$ proposed RoI, or *anchor*, enclosing an object (anchors with ≥0.5 Intersection-over-Union (IoU) overlap with a ground-truth bounding box are considered positive; anchors with <0.5 IoU overlap are considered negative). $p_i^*$ denotes if the anchor is positive ($p_i^* = 1$) or negative ($p_i^* = 0$). Vector $q_i$ represents the four coordinates, characterizing the predicted anchor bounding box, whereas vector $q_i^*$ represents the coordinates for the ground-truth box corresponding to a positive anchor. $L_{cls}$ for each anchor is calculated as log loss for two class labels (object *vs.* non-object). $L_{reg}$ is a regression loss function robust to the outliers, as defined in [20]. $n_{cls}$ and $n_{reg}$ are the normalization parameters for classification and regression losses, respectively, weighted by a parameter $\mu$ [5]. $L_{mask}$ is computed as average cross-entropy loss for per-pixel binary classification, applied to each RoI [6].

**Implementation of *SeBRe*.** The implementation of *SeBRe* generally follows the original work in [6], with limited hyper-parameter optimization for the brain section dataset. Training on the brain section dataset is initialized with pre-trained weights for the MS COCO dataset [10]. Each batch slice consists of a single brain section image per GPU; batch normalization layers are inactivated to optimize training for the small effective batch size. Training is performed using an NVIDIA GeForce GTX 970 GPU. The training regime comprises of two stages. First, the network heads are trained for 6000 iterations, at a learning rate of 0.001 and learning momentum of 0.9. In the second stage, all layers are fine-tuned for 9000 iterations, at a reduced learning rate of 0.0001. During inference, diverging from the original model, the mask branch is applied to the highest scoring eight detection boxes, proposed by the RPN. The maximum number of ground truth instances detected per image is also limited to eight (to avoid erroneous duplicate instances of region-specific masks). Adopting a more stringent approach, the minimum probability threshold for instance detection is raised to 0.9, in order to improve the accuracy of instance segmentation.

**Code availability.**
We provide the code for the *SeBRe* toolbox at https://github.com/itsasimiqbal/SeBRe and https://bitbucket.org/theolab/.

**Data availability.**
The data that support the findings of this study are available from the corresponding author upon reasonable request. The publicly available datasets that are used in this study are available at brain-map.org/api/index.html and https://www.nitrc.org/frs/shownotes.php?release_id=2316. The annotated datasets that are used in this study are available at https://github.com/itsasimiqbal/SeBRe and https://bitbucket.org/theolab/.

| Brain Region | *SeBRe* Hausdorff distance (mean) | F-CNN Hausdorff distance (mean) | *SeBRe* CMD (mean) | F-CNN CMD (mean) | *SeBRe* DICE (mean) | F-CNN DICE (mean) | BSegNet DICE (mean) | QNAT DICE (mean) |
|---|---|---|---|---|---|---|---|---|
| Caudate (Left) | **1.874** | 6.5 | **0.374** | 0.75 | **0.886** | 0.785 | 0.86 | 0.875 |
| Pallidum (Left) | **1.971** | 4.5 | **0.477** | 0.77 | **0.849** | 0.77 | 0.81 | 0.81 |
| Putamen (Left) | **1.828** | 5.0 | **0.412** | 0.72 | 0.889 | 0.84 | **0.91** | 0.89 |
| Thalamus (Left) | **2.44** | 5.0 | **0.557** | 0.77 | **0.883** | 0.88 | 0.88 | 0.88 |
| Caudate (Right) | **2.044** | 6.5 | **0.386** | 0.75 | **0.884** | 0.8 | 0.88 | 0.87 |
| Pallidum (Right) | **2.124** | 4.5 | **0.502** | 0.80 | **0.844** | 0.75 | 0.83 | 0.825 |
| Putamen (Right) | **1.875** | 7.0 | **0.449** | 0.75 | 0.872 | 0.85 | **0.91** | 0.89 |
| Thalamus (Right) | **2.206** | 4.5 | **0.518** | 0.75 | 0.891 | 0.885 | **0.9** | 0.878 |

**TABLE 1.** Performance comparison of *SeBRe* with other deep learning-based brain region segmentation techniques. Highlighted scores represent the best performance of a DNN on the given brain region.

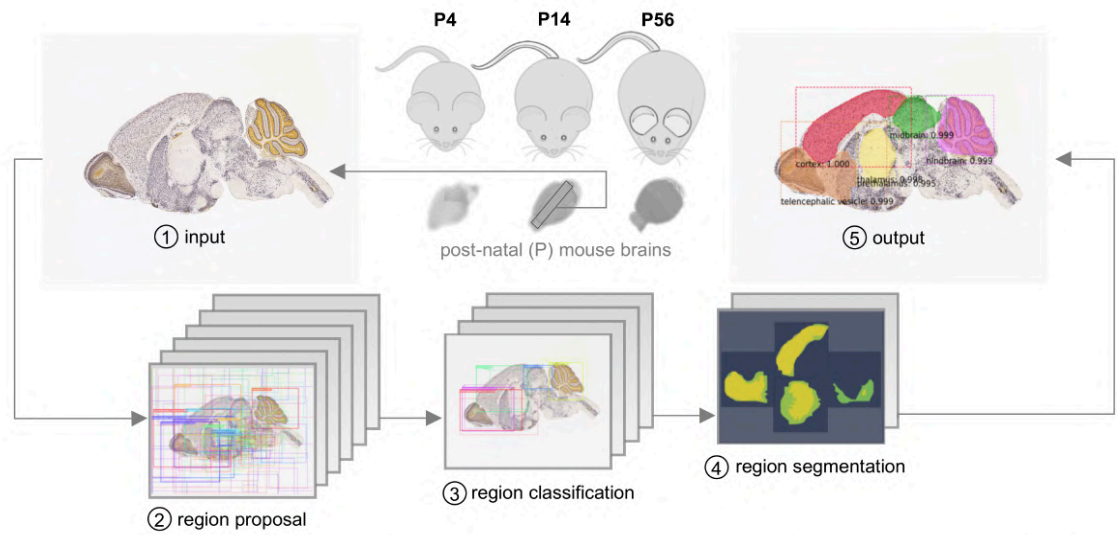

**Fig. 1 |** *SeBRe* **(deep learning pipeline) block diagram architecture**. Medial section (top left) of a P14 GAD1 mouse brain is fed as input to the network and the output (top right) shows the segmented brain regions on top of the input brain section, after brain region proposal, classification and segmentation stages.

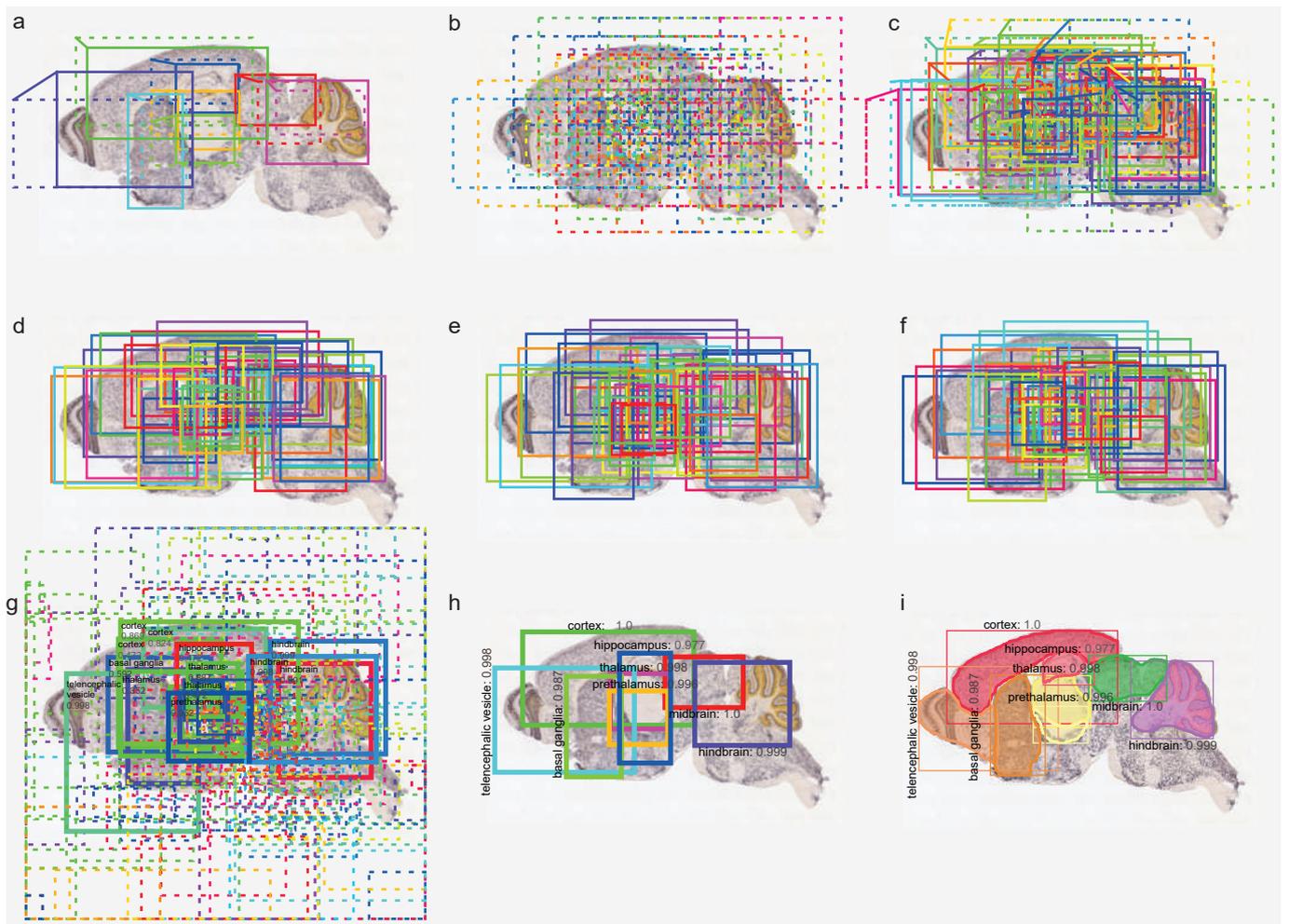

**Fig. 2 | *SeBRe* multistage image processing pipeline.** Step-by-step flow of brain region localization (a-f), classification (g-h) and segmentation (i): **a**, the bounding boxes of ground-truth targets, generated for Region Proposal Network (RPN) training. **b**, boxes for proposed regions (anchors) that are predicted by the RPN. **c**, the RPN-predicted anchors after refinement. **d**, the RPN-predicted anchors after clipping to the image boundaries. **e**, the RPN-predicted anchors after applying non--maximum suppression. **f**, final RPN-predicted anchors are shown after coordinate normalization. **g**, predictions of the Feature Pyramid Network (FPN) classifier heads, on RPN-predicted anchors before refinement are shown. Filled boxes are labelled with predicted class and confidence score; dashed boxes represent proposals classified as background. **h**, FPN-predicted classes are shown with their classification scores after anchor refinement and non-maximum suppression. **i**, final segmented masks of brain regions with their classification scores are shown overlaid on the input brain section.

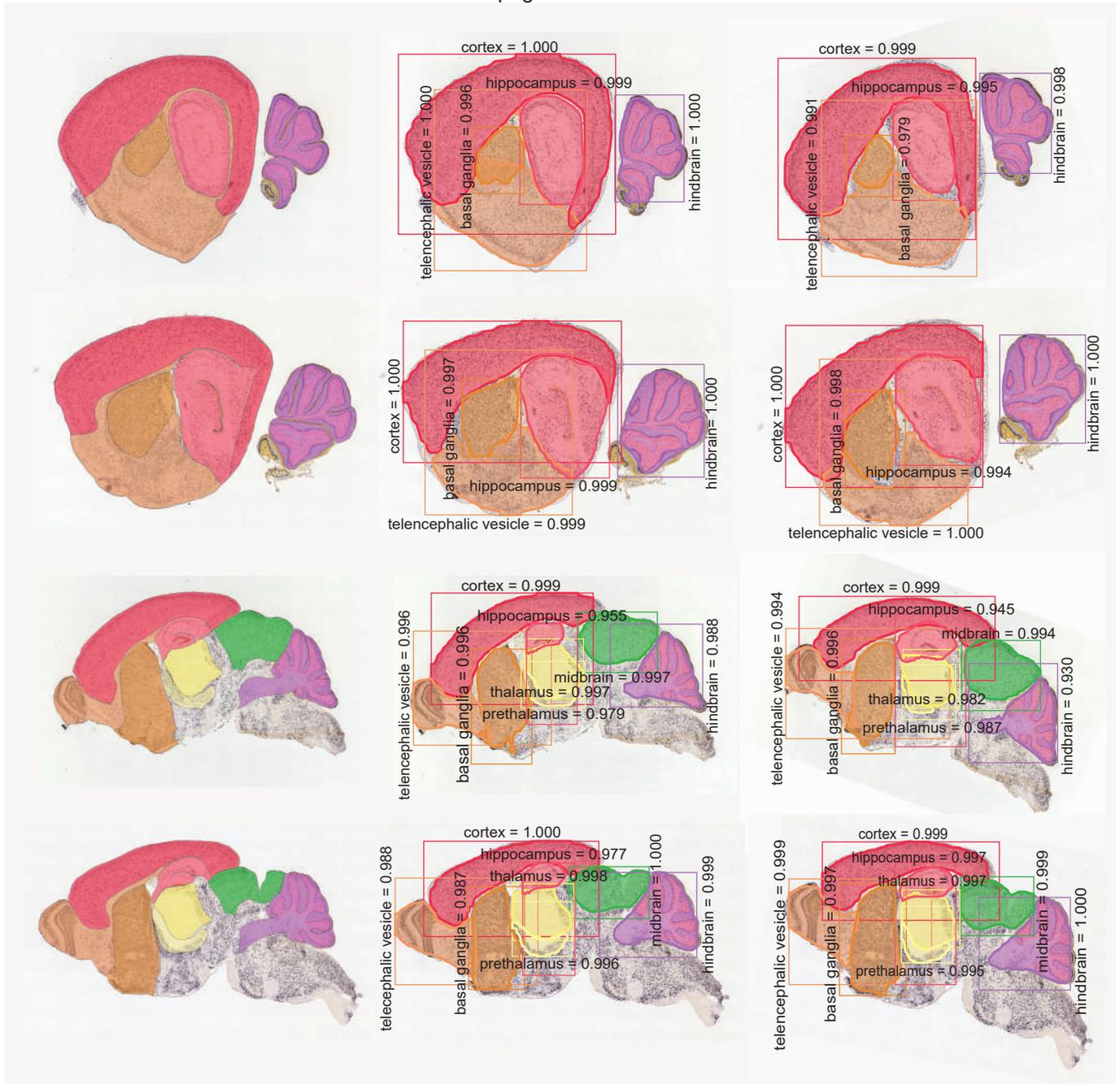

**Fig. 3 | Performance of *SeBRe* in segmenting brain regions.** Qualitative performance comparison of SeBRe on lateral (rows: 1-2) and medial (rows: 3-4) brain sections of P14 mouse brains with human-annotated masks. *SeBRe* performs optimally on predicting masks of brain regions, for both upright (column: 2) and rotated (column: 3) versions of the input brain sections (column: 1).

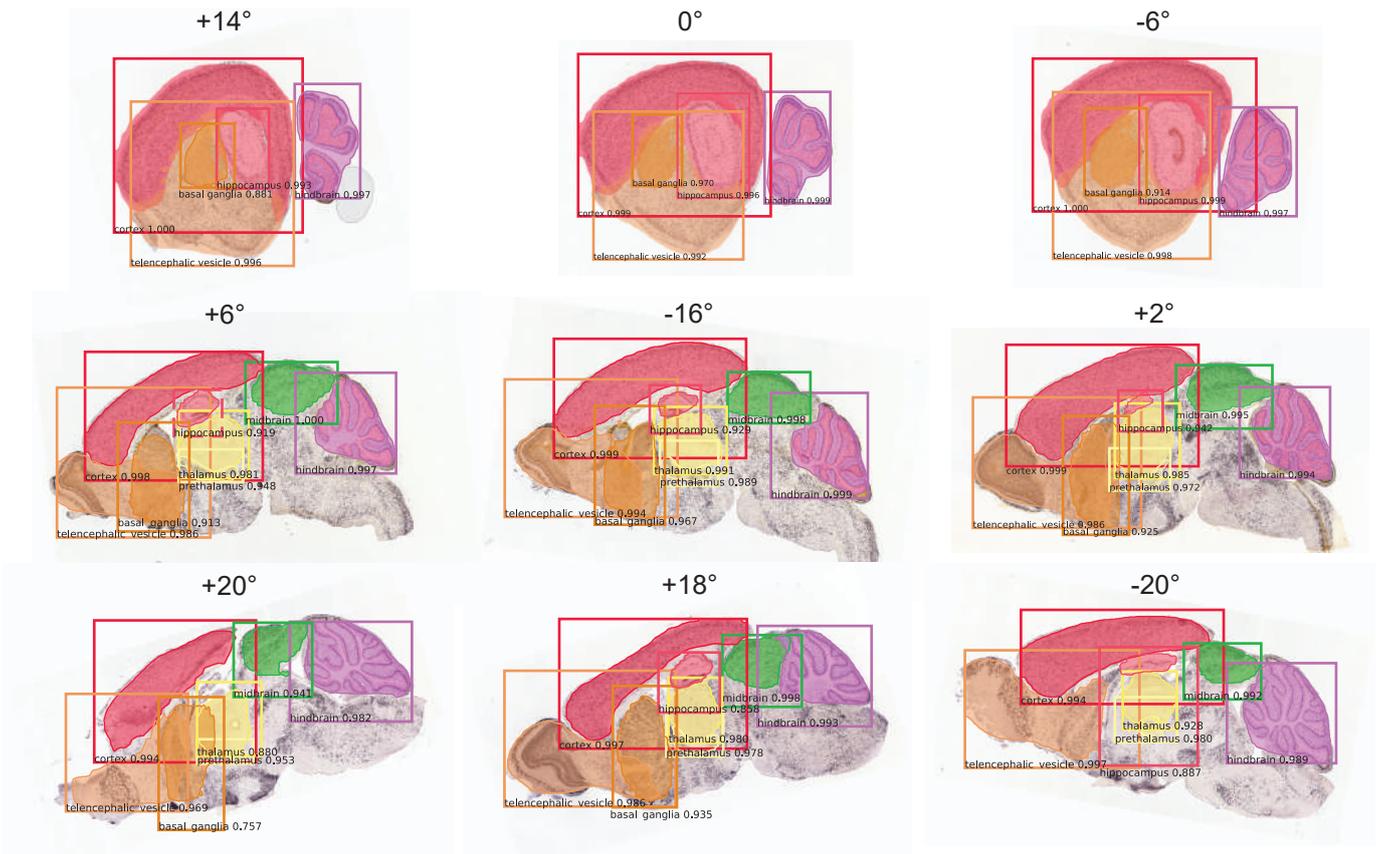

**Fig. 4 | Rotationally-invariant performance of *SeBRe* on the extended mouse dataset.** Performance of *SeBRe* is shown on a set of sample rotated sections of the mouse brain testing dataset. Each rotation is given on top of the brain section, with reference to 0° (upright). *SeBRe*'s performance is invariant to the rotation of a brain section in segmenting brain regions of interest even in cases when the tissue is broken, e.g. cortex and telencephalic vesicle in the bottom-left section.

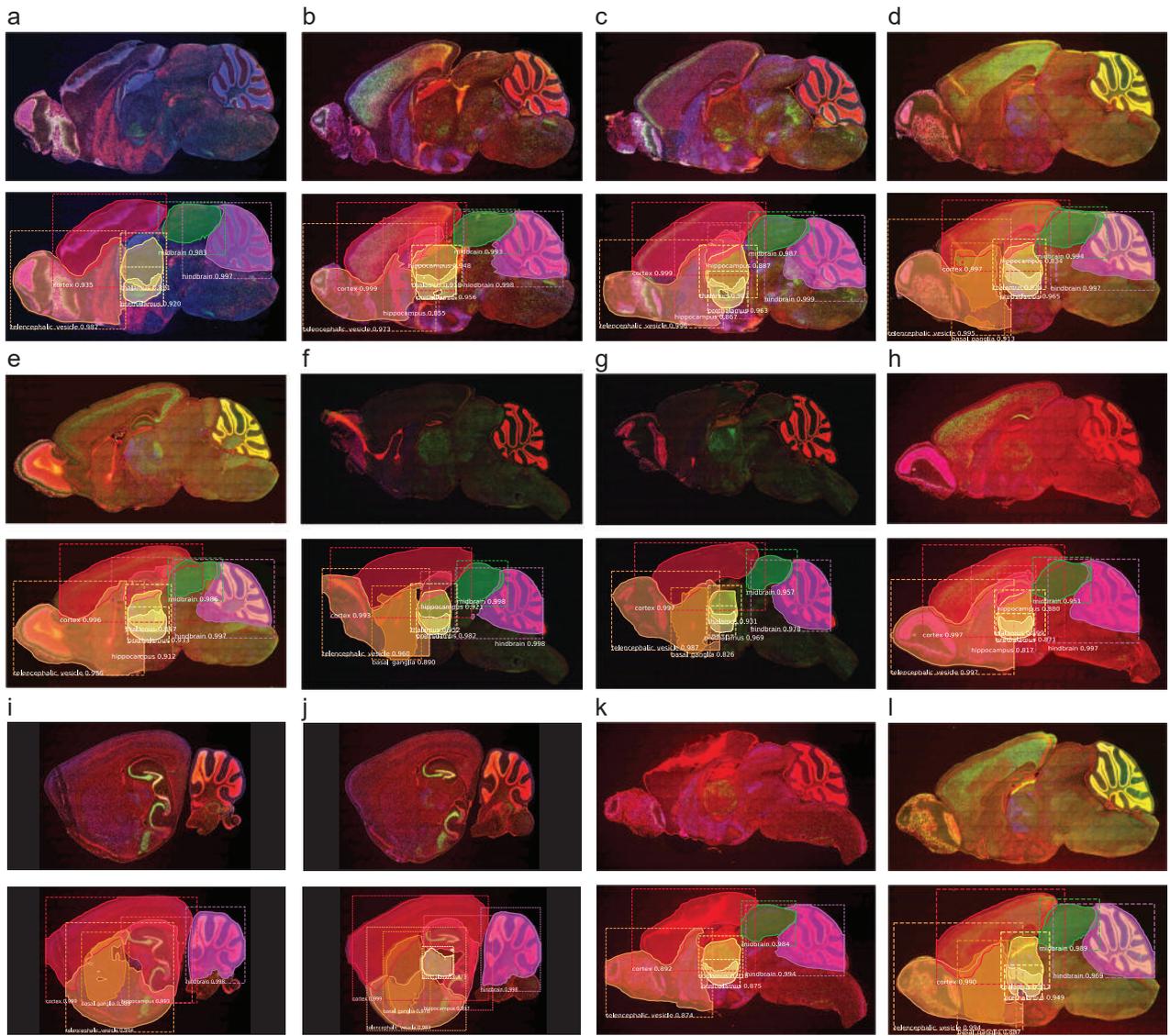

**Fig. 5 | Generalised performance of *SeBRe* on 'unseen' FISH brain sections.** *SeBRe* is trained on the extended mouse brain dataset of ISH brain sections and tested on Fluorescence in situ hybridisation (FISH) brain sections. Input brain sections are shown at the top of each panel, with the output of the network displayed below. *SeBRe* is able to detect the brain regions captured through a different imaging modality without any additional preprocessing. Performance is shown on a variety of FISH brain sections collected randomly from a list of genetic labeling markers available on the Allen Brain Institute, such as GAD1/Cux2-CreERT2- Ai14(tdTomato) (**a-c**), GAD1/Grik4-Cre-Ai14(tdTomato) (**i-j**), Rorb/Scnn1a-Cre-Ai14(tdTomato) (**d,e,l**), Slc17a6/Slc32a1-Cre-Ai14(tdTomato) (**f,g**) and GAD1/Gpr26-CreKO250- Ai14(tdTomato) (**h,k**).

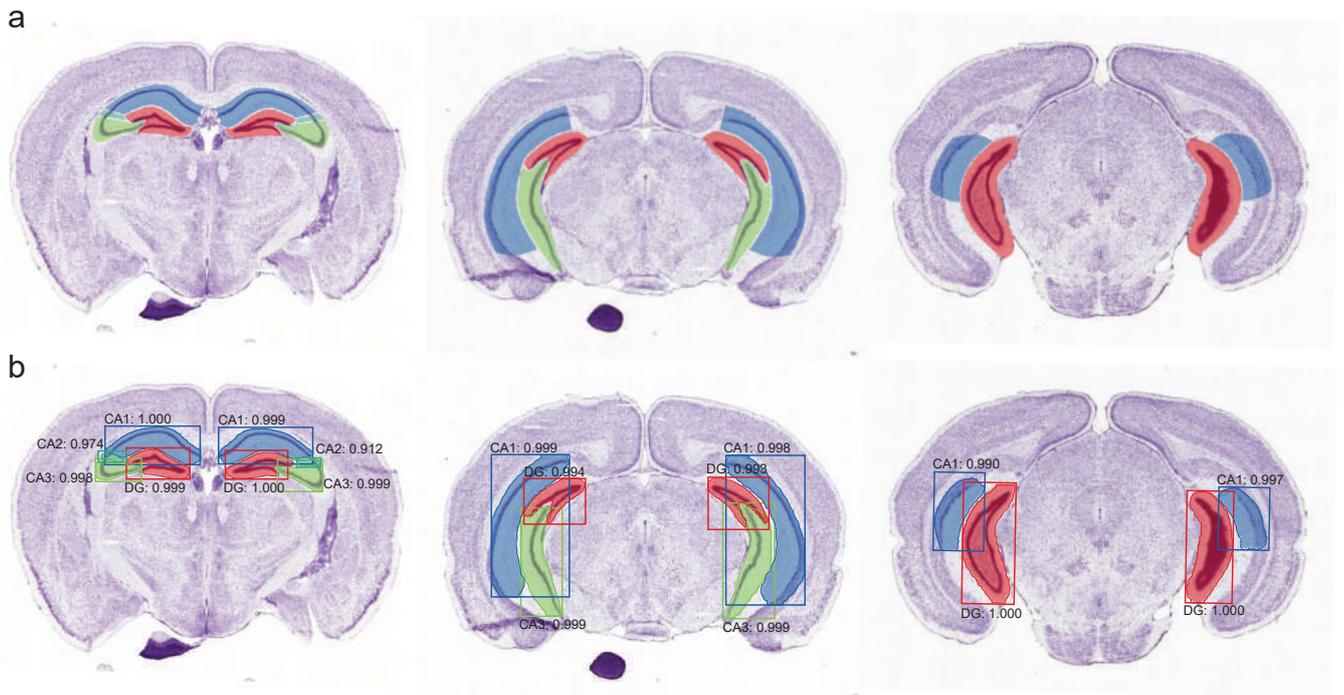

**Fig. 6 | Performance of *SeBRe* in segmenting sub-regions of hippocampus. a**, Three mouse brain sections taken randomly from different coronal planes, with overlaid human-annotated (ground-truth) masks for hippocampal sub-regions (CA1, CA2, CA3 and DG) are shown. **b**, Performance of *SeBRe* is demonstrated in segmenting hippocampal sub-regions in the corresponding upright brain sections in (a).

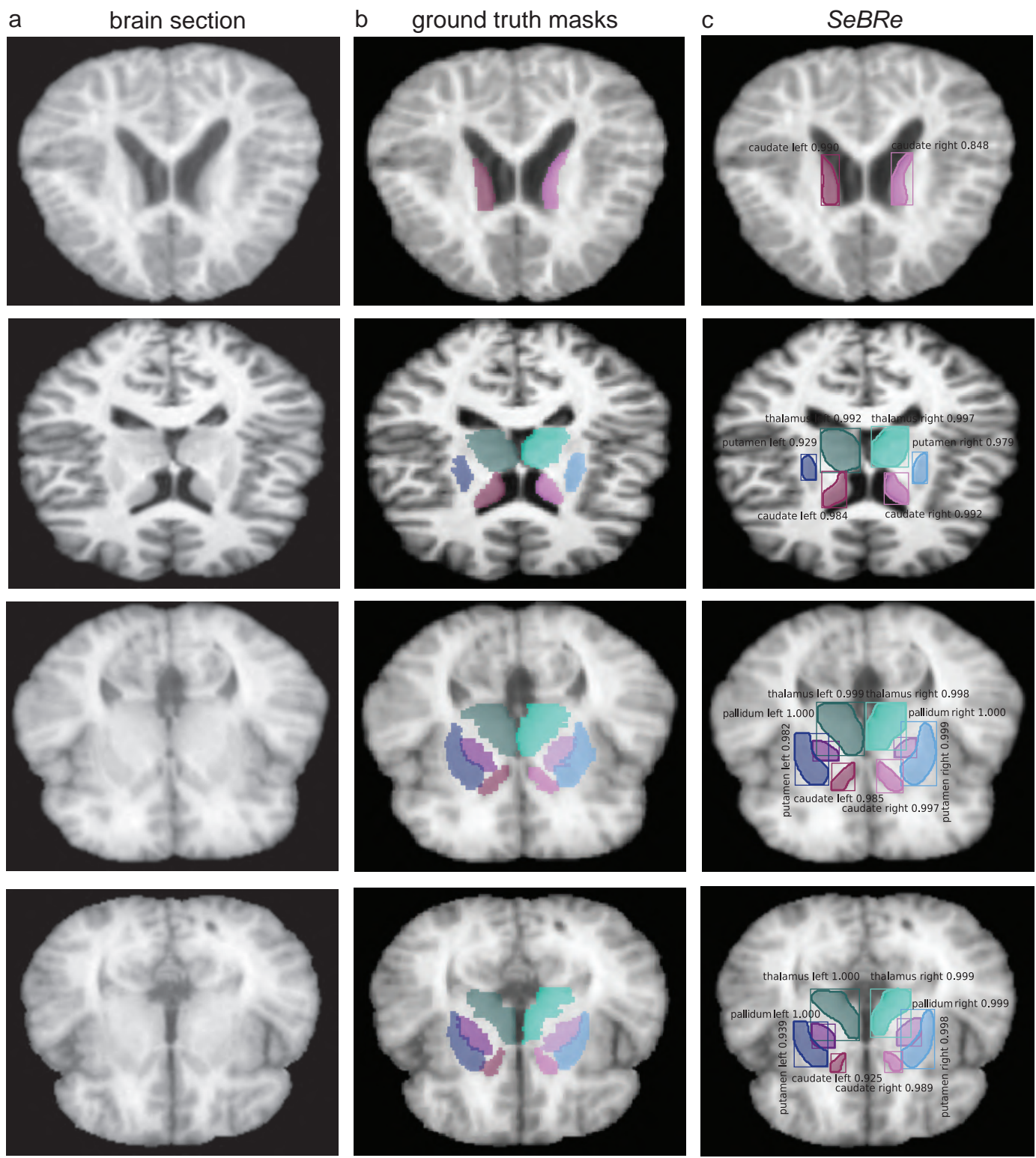

**Fig. 7 | Performance of *SeBRe* on human brain sections. a**, Randomly selected human brain MR scans are shown. **b**, Ground-truth masks of the corresponding MR scans in (a). **c**, Performance of *SeBRe* is demonstrated in detecting eight brain regions: caudate (left), caudate (right), thalamus (left), thalamus (right), putamen (left), putamen (right), pallidum (left) and pallidum (right) in the human MR scans.

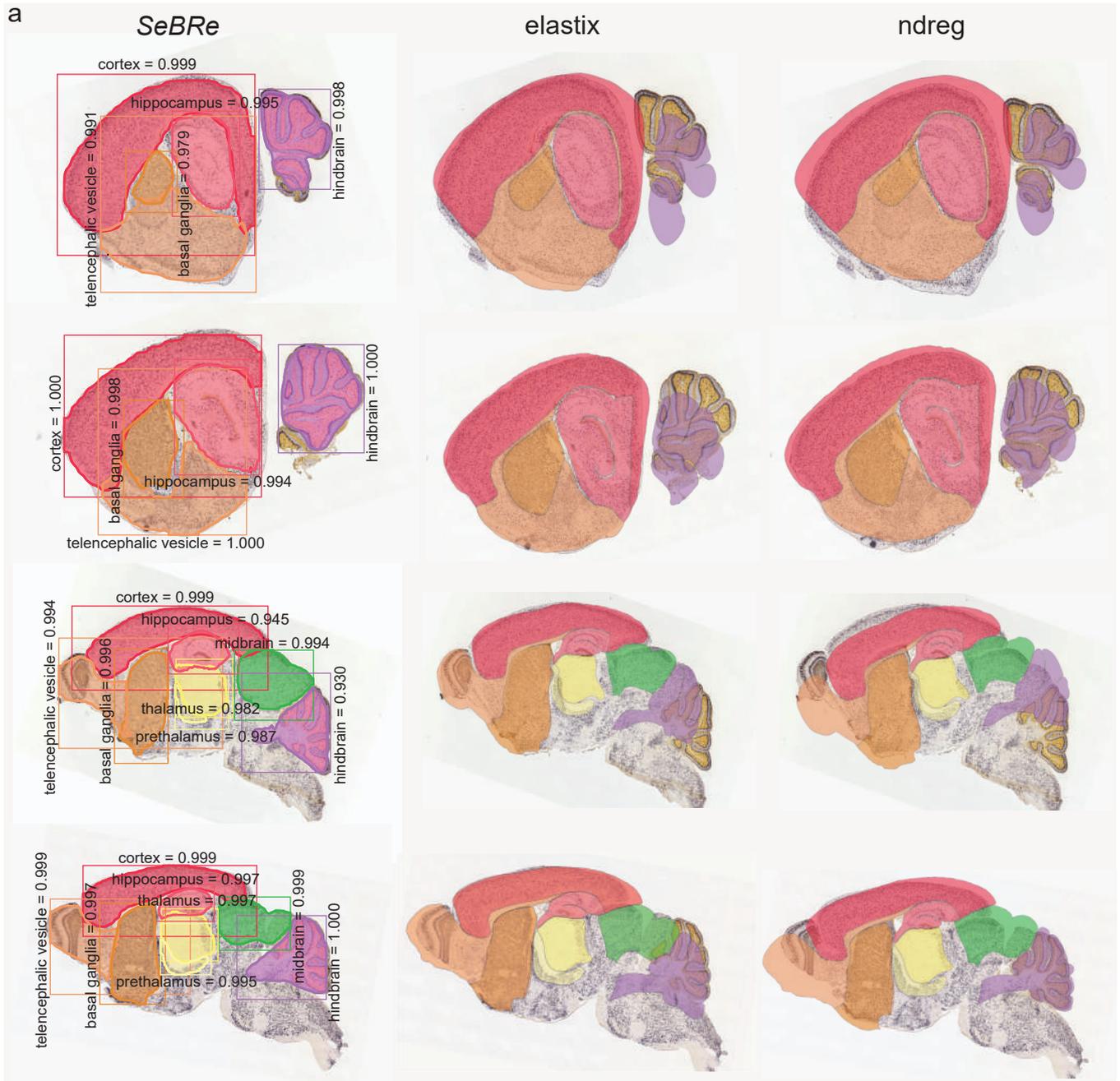
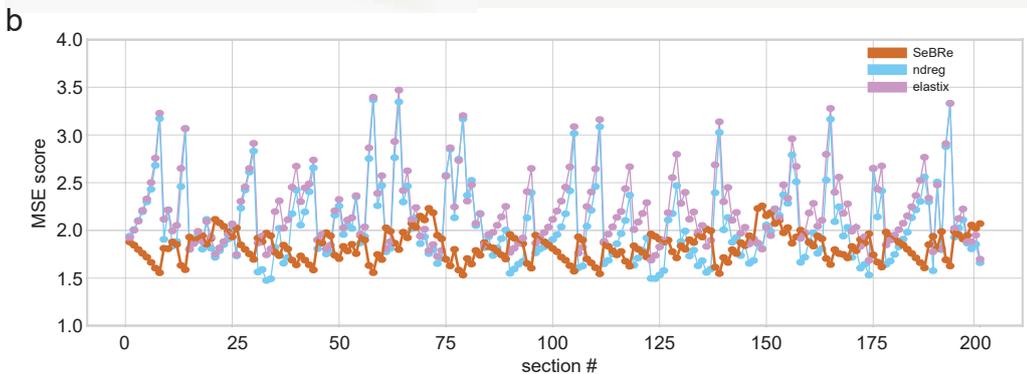
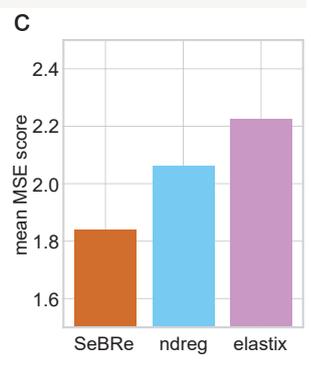

**Fig. 8 | Comparision of *SeBRe* with commonly used brain registration methods. a**, Performance of *SeBRe* is demonstrated on randomly selected lateral and medial brain sections in comparision with commonly used brain registration methods, ndreg and elastix. **b**, Plot of MSE scores for all brain sections in the test dataset, for *SeBRe*, ndreg and elastix. **c**, Mean MSE scores for *SeBRe*, ndreg and elastix on the complete test dataset.

| **a** brain section before annotation | **d** brain section after human annotation |

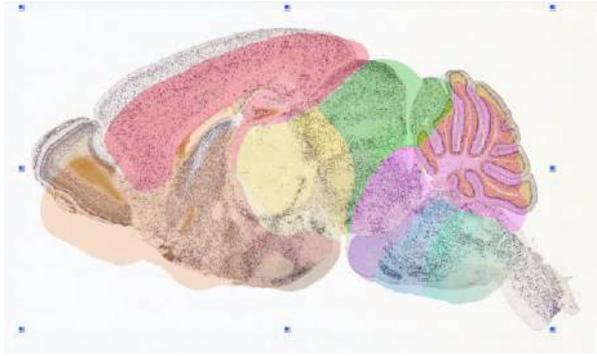 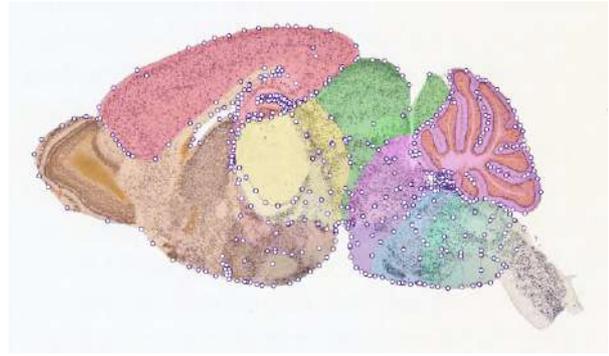

| **b** cortex annotation | before vector alignment | after vector alignment |

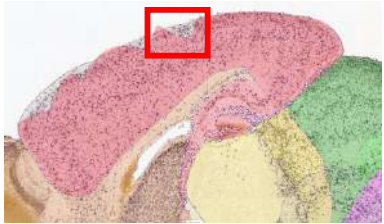 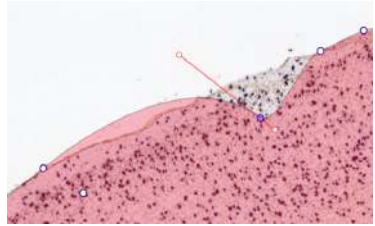 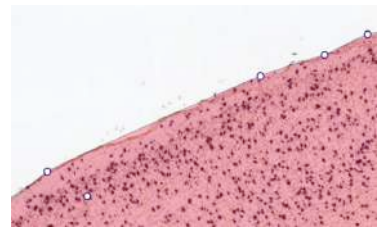

| **c** hindbrain annotation | before vector alignment | after vector alignment |

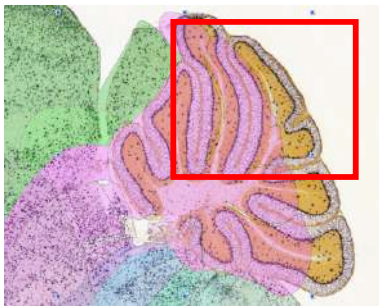 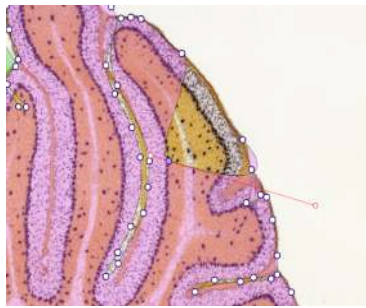 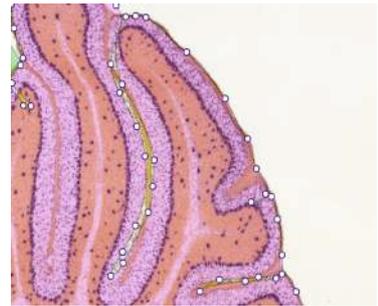

**Supplementary Fig. 1 | Human expert annotation to generate ground-truth masks of mouse brain sections dataset. a**, Medial brain section of P14 mouse brain is shown with a non-registered reference atlas on top. **b-c**, Step by step annotation of cortex and hindbrain regions with boundary alignment is presented. Boundary points are handled by carefully aligning scalable vectors by using a Scalable Vector Graphics (SVG) software. **d**, Brain section after human annotation is presented.

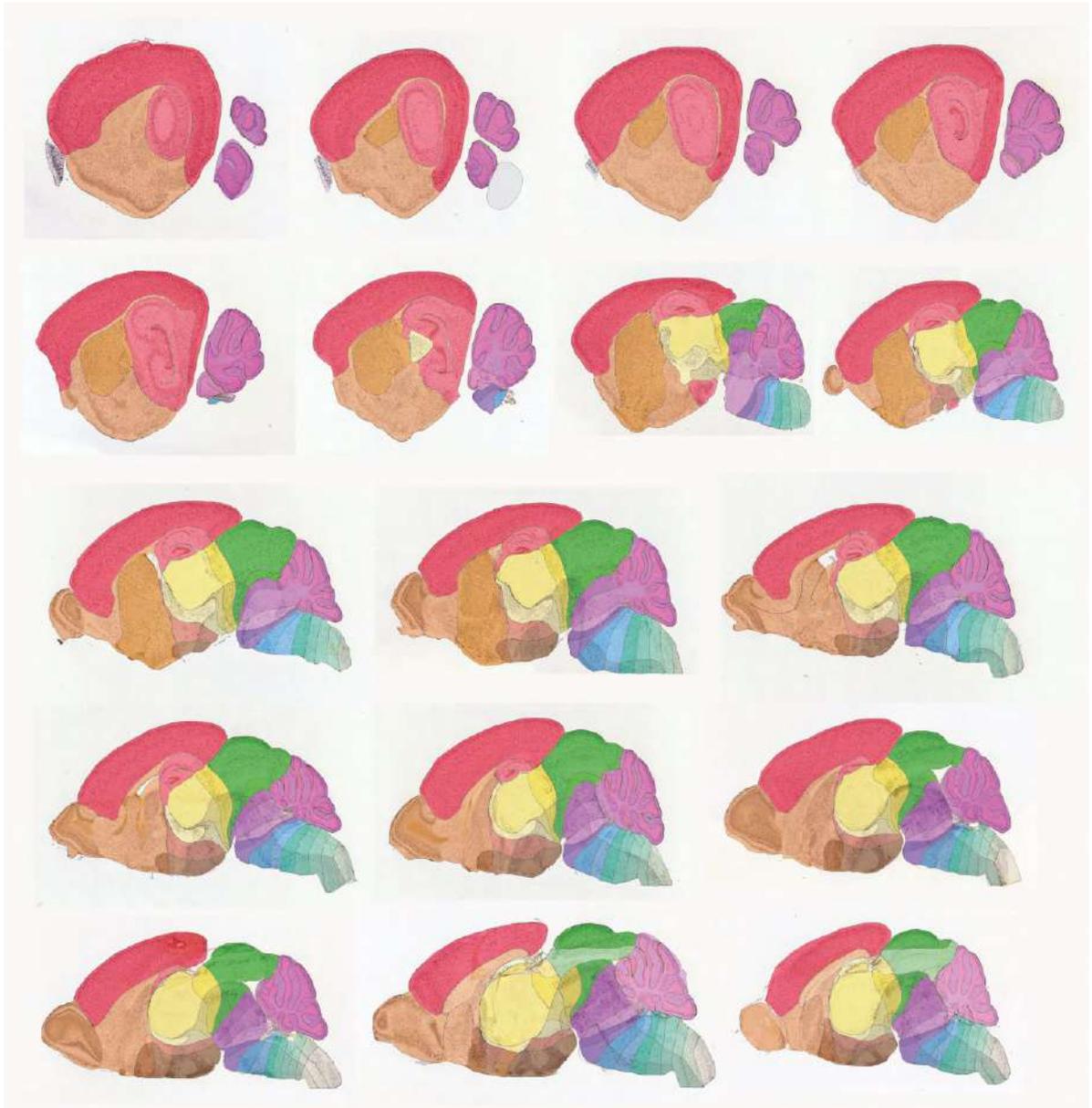

**Supplementary Fig. 2 | Manually-registered ground-truth data.** P14 GAD1 sagittal sections with overlaid Allen developing mouse brain reference atlas are shown: lateral-most (top left) to medial-most (bottom right).

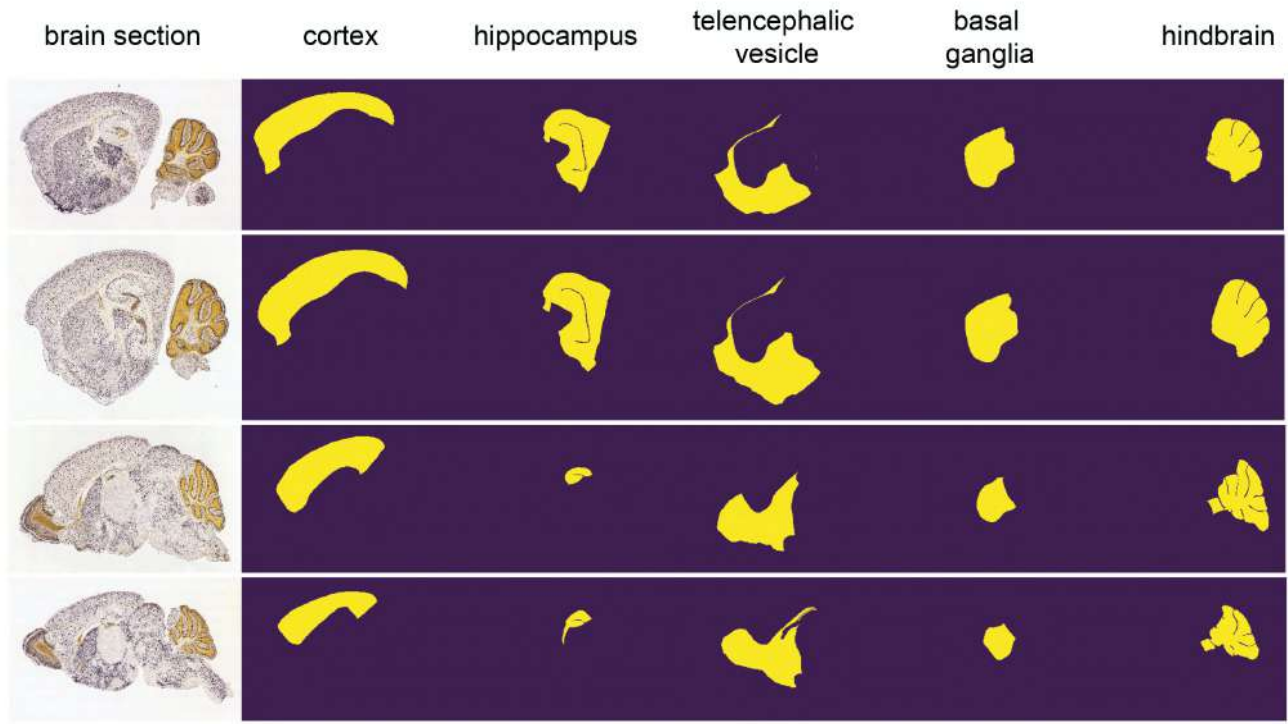

**Supplementary Fig. 3 | Masks of brain regions on lateral and medial sagittal sections of P14 GAD1 and VGAT ground-truth mouse brains.** The first two rows of column 1 show the lateral, whereas the last two rows show the medial brain sections. Column 2-6 shows the ground-truth masks of five example brain regions. These regions vary in shape and size as we move from lateral to medial (e.g. compare hippocampus across column 3, for all brain sections).

a P4 GAD1

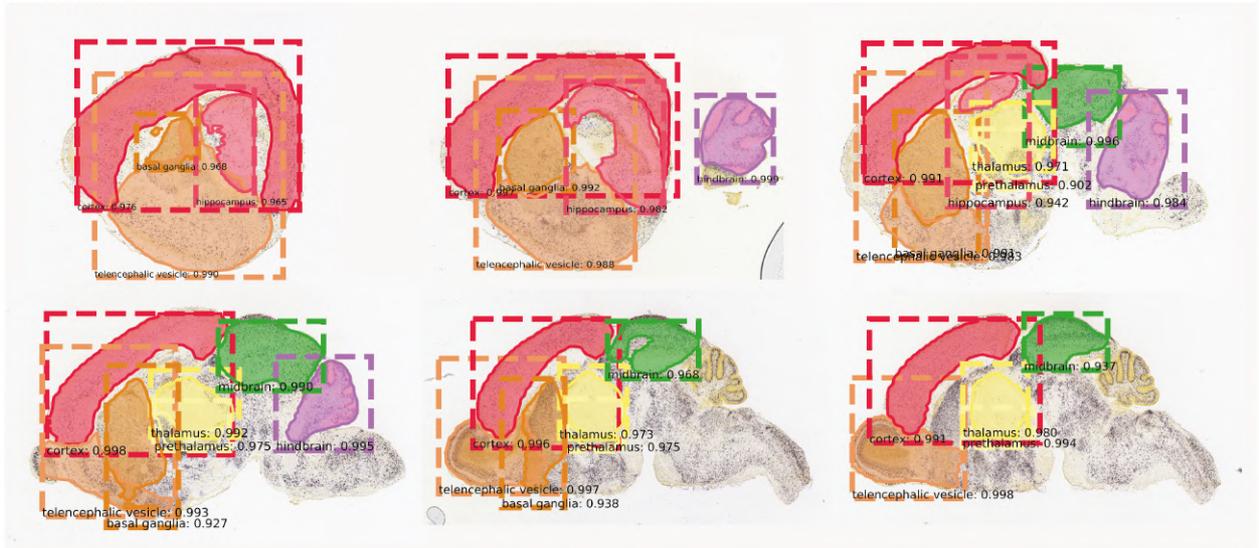

b P4 VGAT

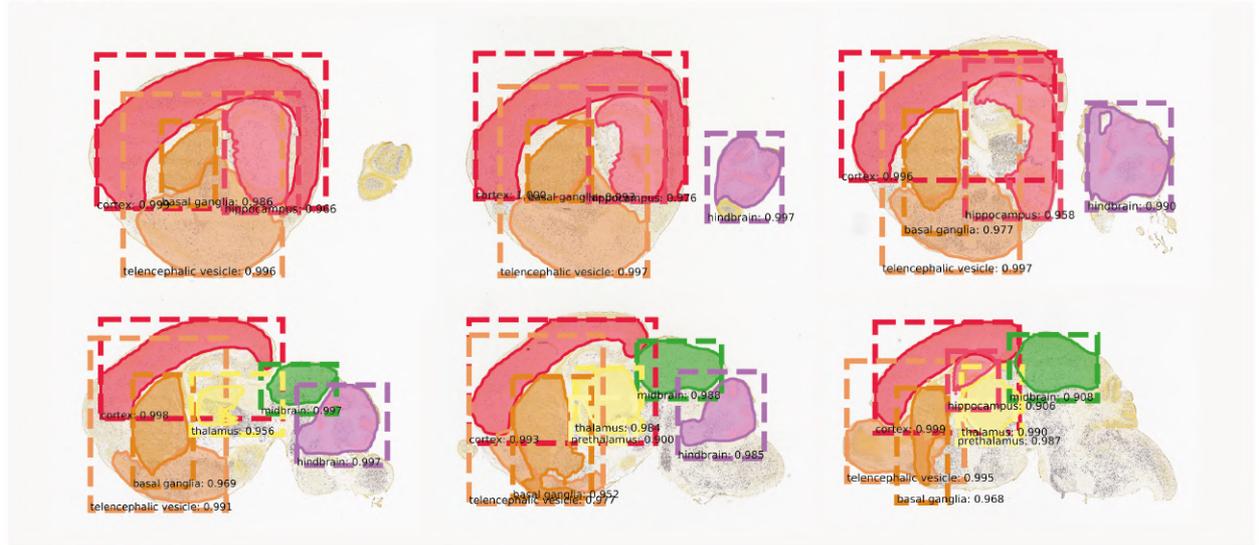

c P4 Nissl

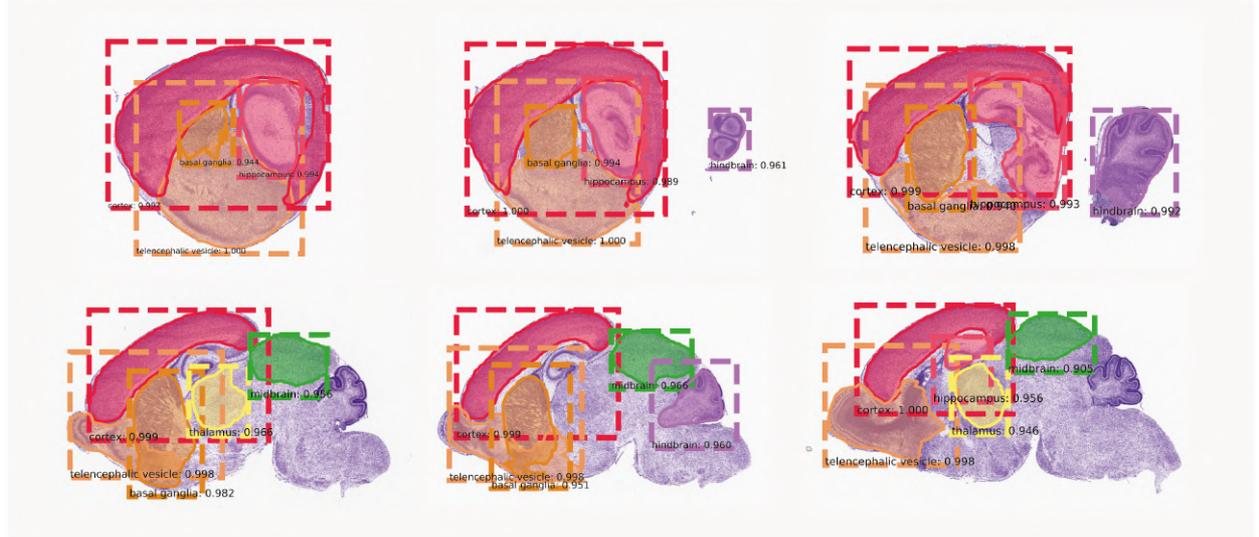

## d P4 CamKII

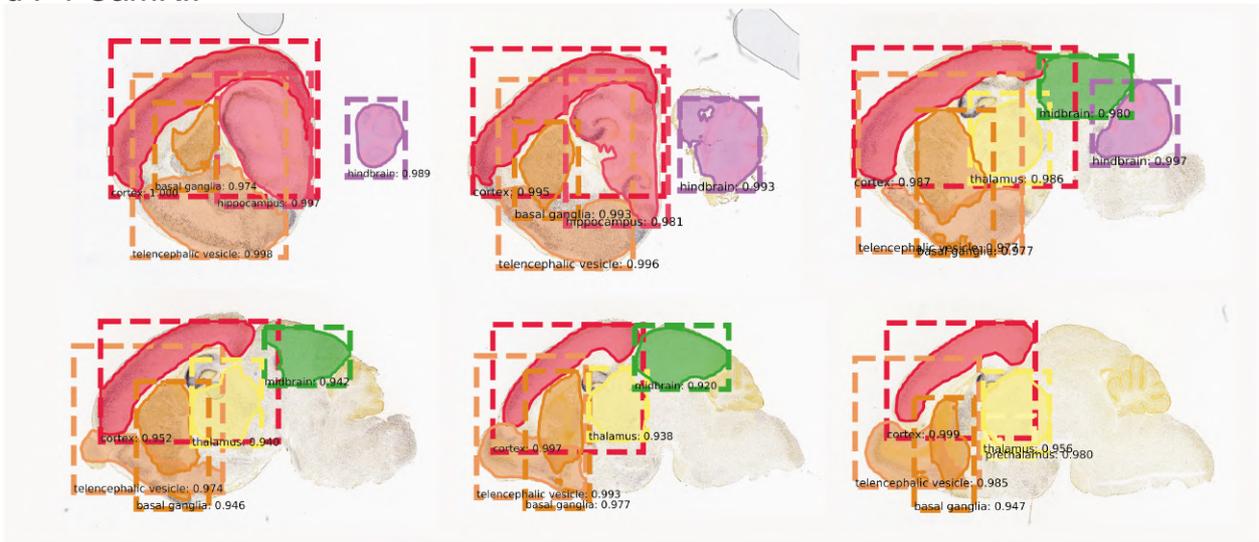

## e P14 Nissl

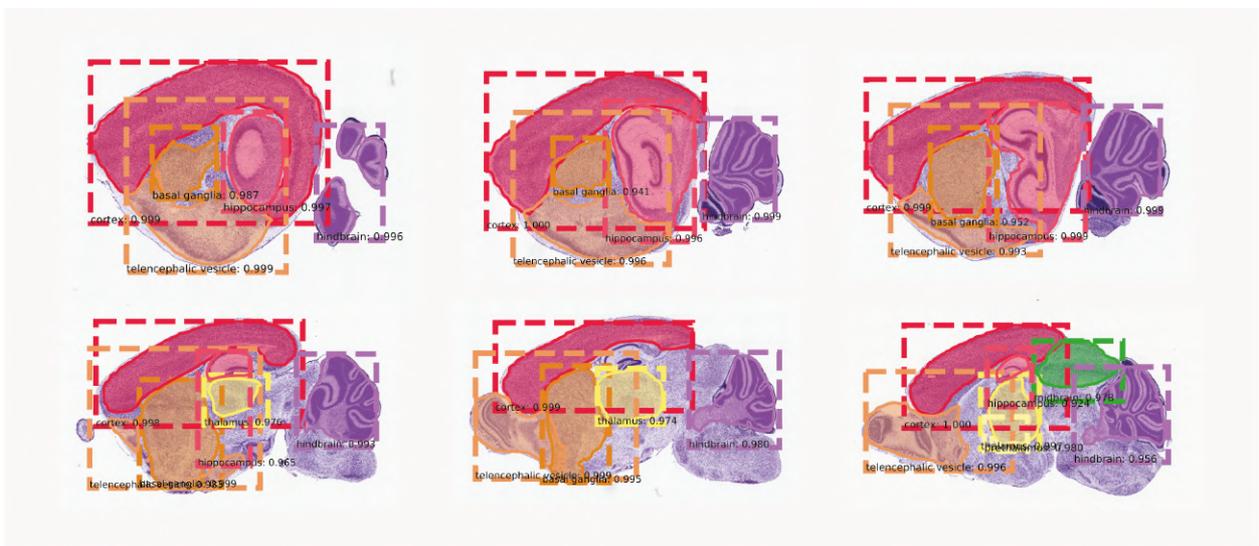

## f P14 CamKII

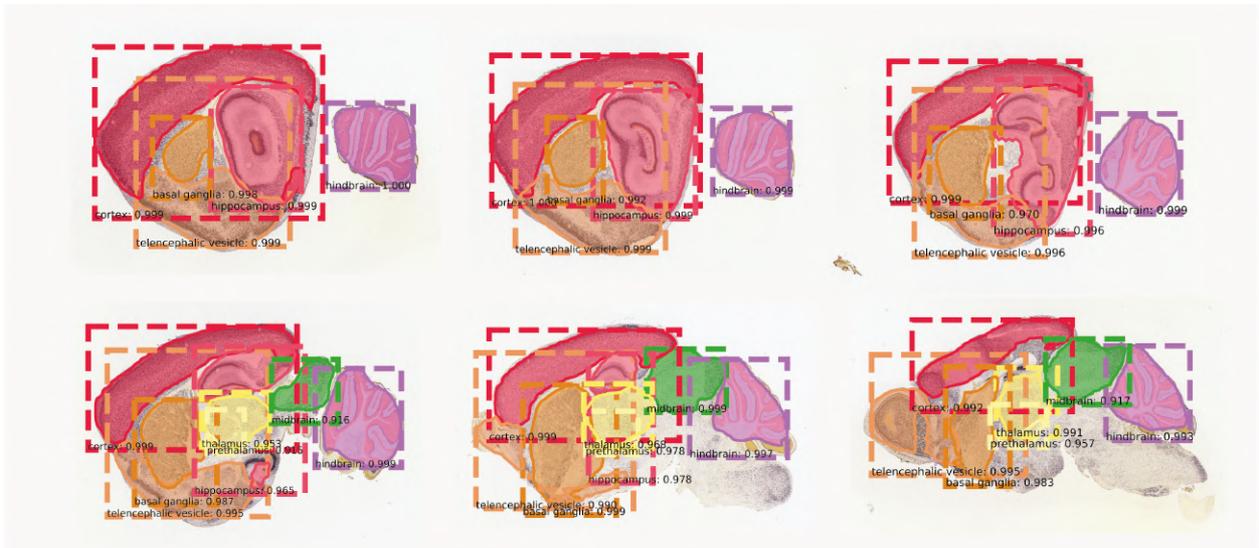

## g P28 GAD1

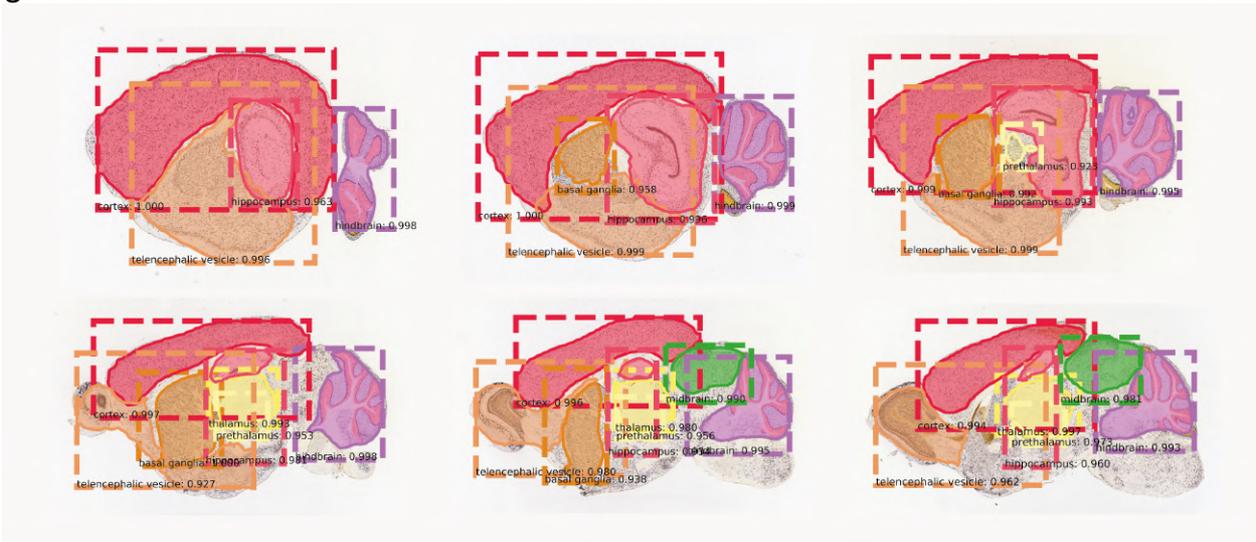

## h P28 VGAT

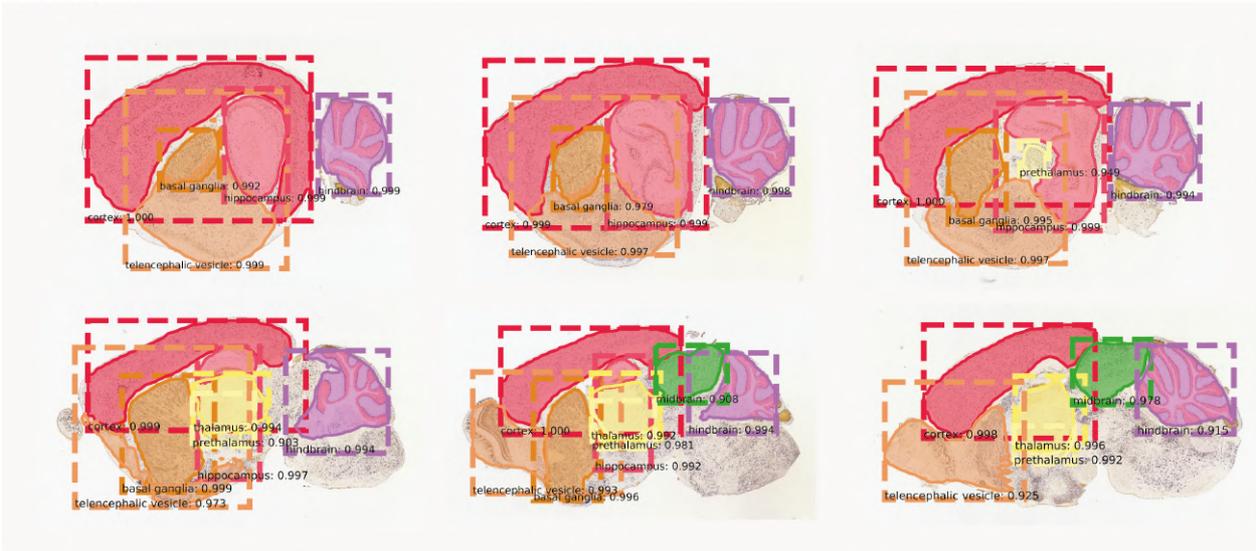

## i P28 CamKII

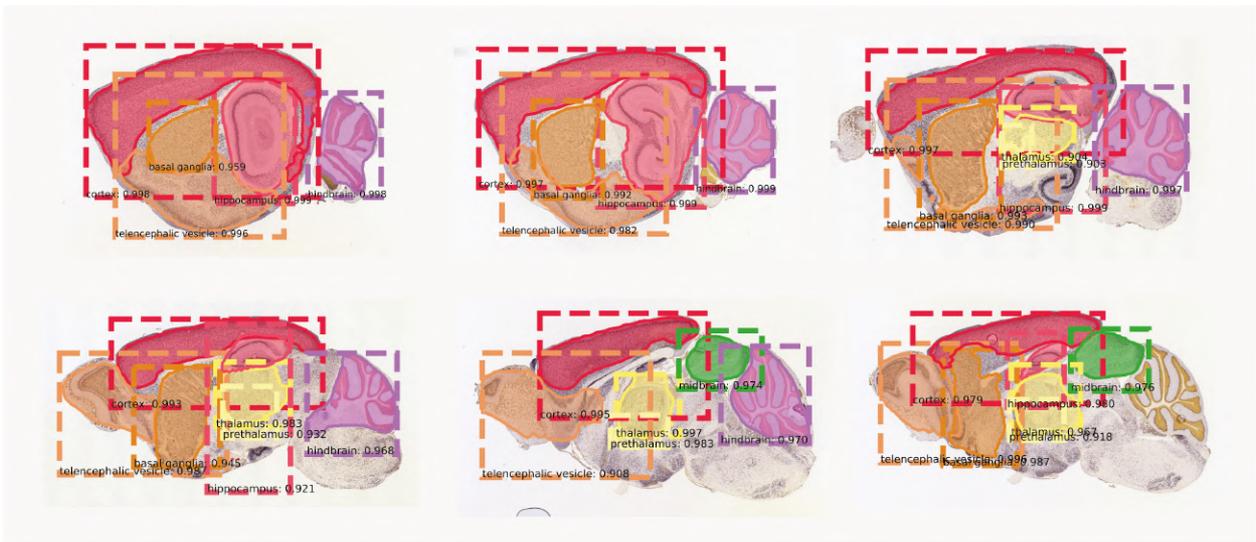

## j P56 Nissl

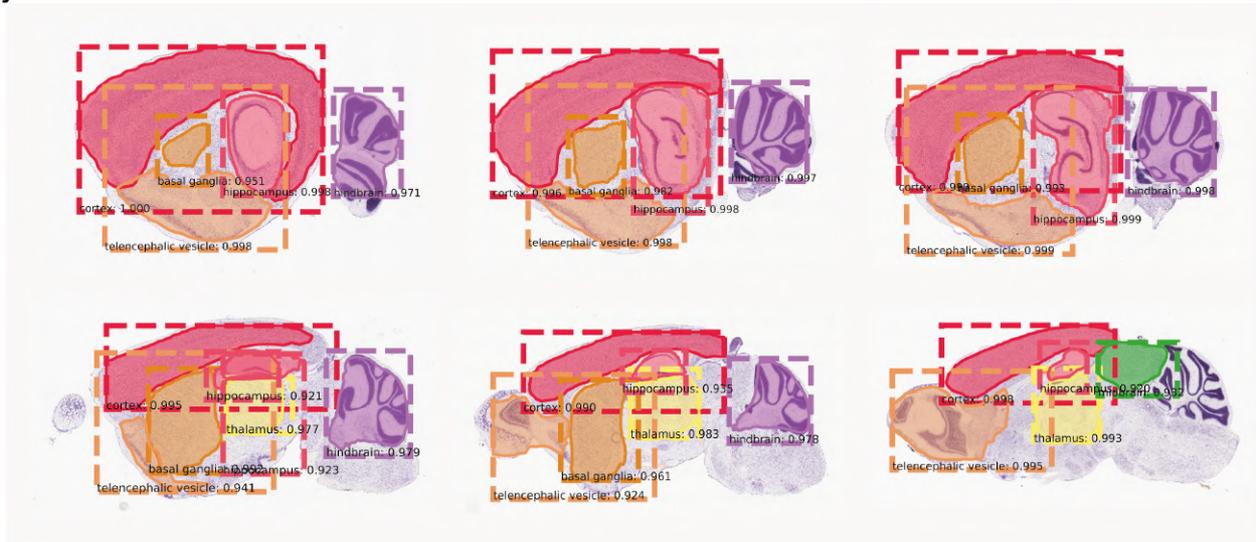

**Supplementary Fig. 4 | Performance of *SeBRe* on various 'unseen' mouse brain ISH tissues, across different developmental time points. a**, Predicting masks of brain regions by *SeBRe* on GAD1 brain sections at age P4. **b**, on VGAT at age P4, **c**, on Nissl at age P4, **d**, on CamKII at age P4, **e**, on Nissl at age P14, **f**, on CamKII at age P14, **g**, on GAD1 at age P28, **h**, on VGAT at age P28, **i**, on CamKII at age P28 and **j**, on Nissl at age P56.

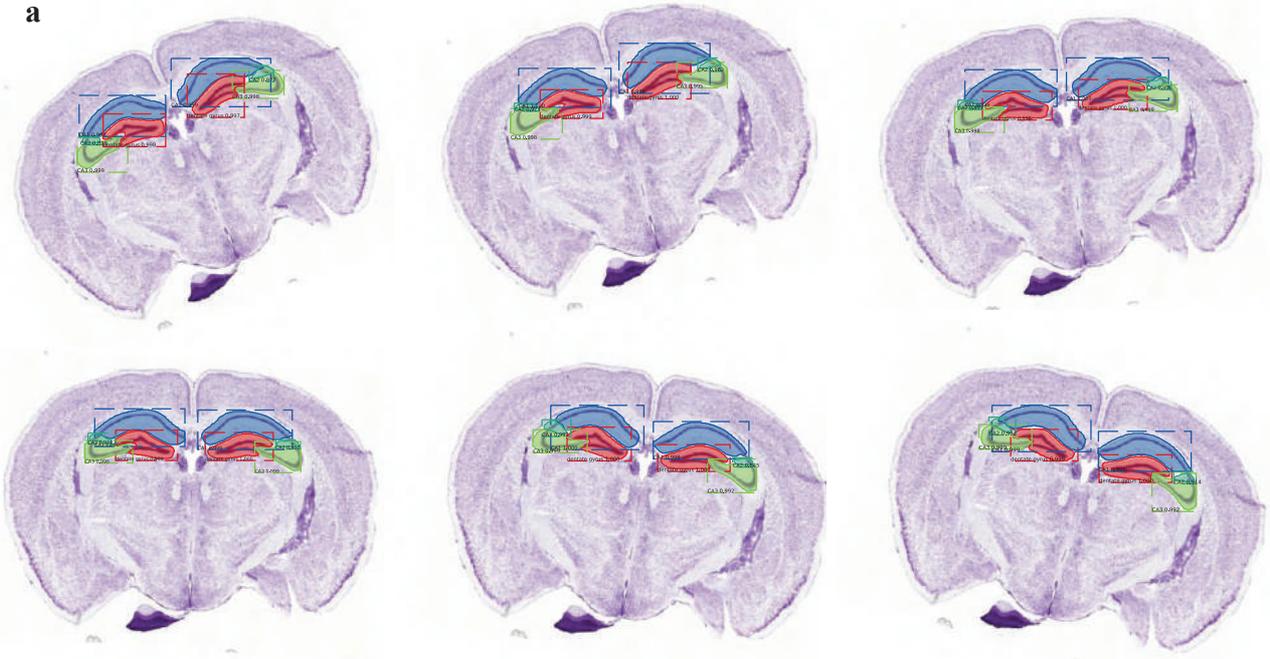

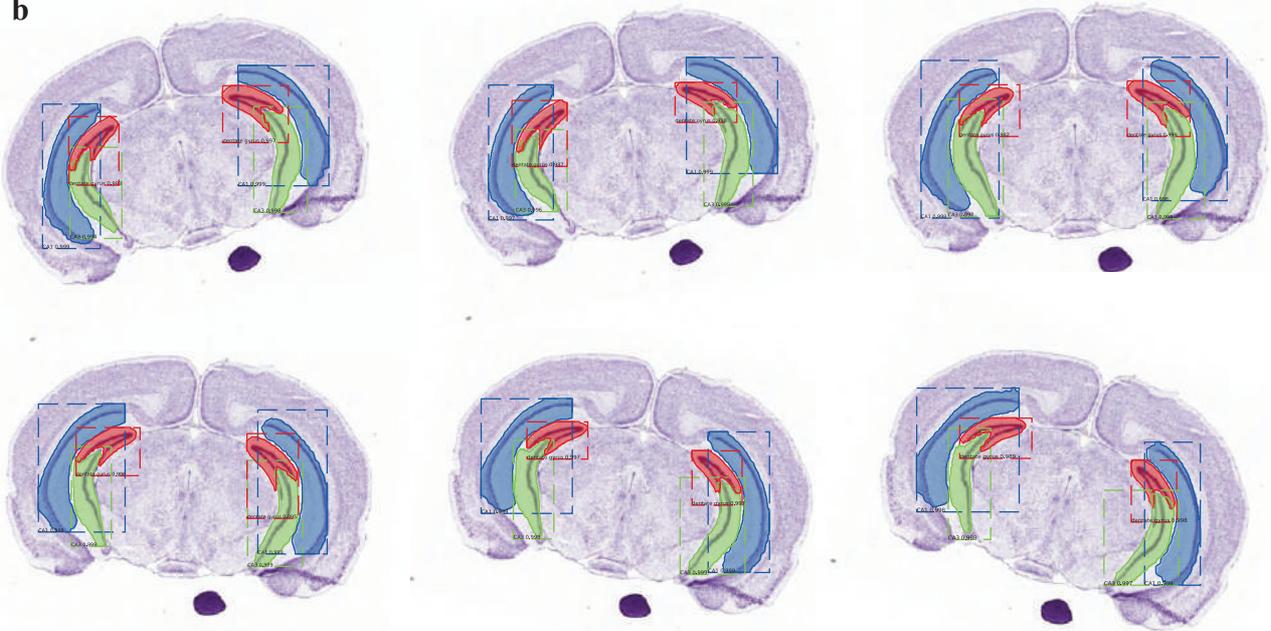

**Supplementary Fig. 5 | Rotationally-invariant performance of *SeBRe* on hippocampus sub-regions's dataset.** Performance of SeBRe is shown on two sets of rotated sections of mouse brain testing dataset, one from the rostral (**a**) and the other from the caudal (**b**) plane. Sections are rotated from -20 to +20 degrees with reference to 0° (upright). *SeBRe*'s performance is insensitive to the rotation of a brain section in segmenting brain sub-regions of interest.

a 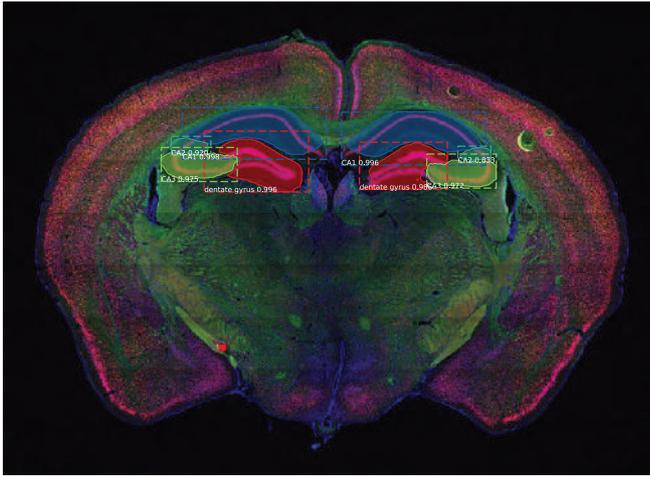 b 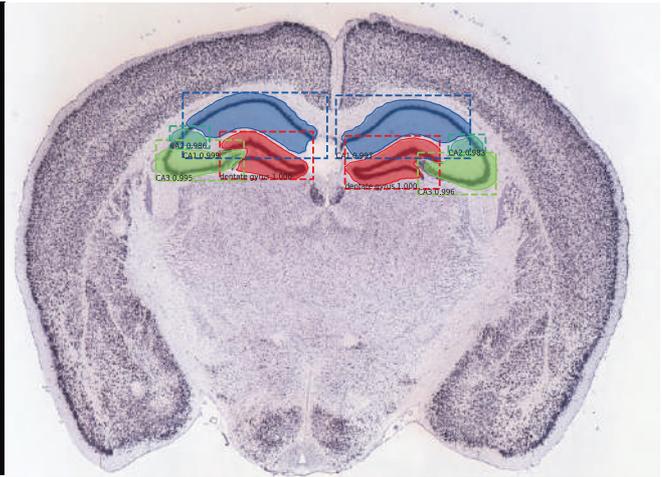

**Supplementary Fig. 6 | Performance of SeBRe in segmenting sub-regions of hippocampus in different imaging modality and genetic marker. a**, Performance of *SeBRe* in detecting hippocampal sub-regions in FISH brain section from NeuN (NF-160) marker is shown. **b**, Performance of SeBRe in detecting hippocampal sub-regions in a randomly ISH brain section from Nrgn marker is shown.

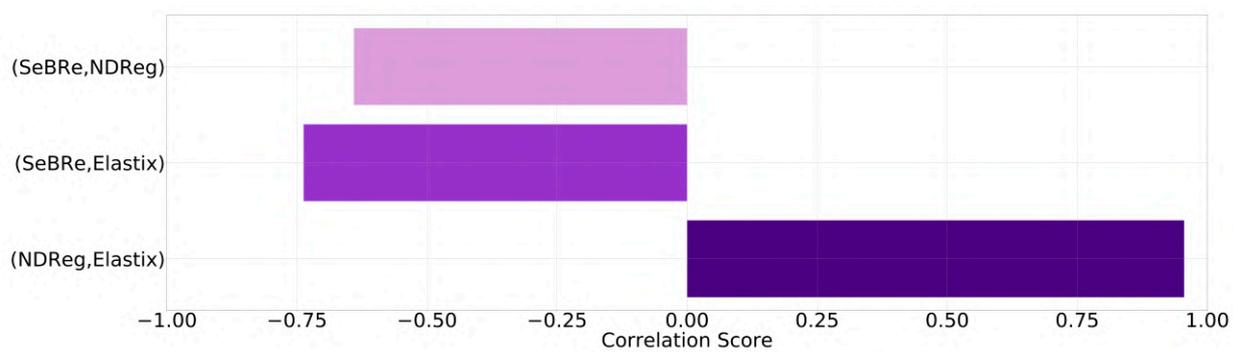

**Supplementary Fig. 7 | Pairwise correlation of *SeBRe*, ndreg and elastix MSE scores on testing dataset.** It appears that ndreg and elastix have a strong positive correlation in their MSE scores on the complete testing dataset of brain regions. However, interestingly, *SeBRe* is negatively correlated with these methods. The plot points towards the fact that the drop in the performance of *SeBRe*, which mainly occurs due to the brain regions omitted by the network, is very different from the errors made by ndreg and elastix in registering the brain sections. Since, ndreg and elastix utilize affine and non-affine transformation algorithms and therefore cannot minimize error by omitting regions, they instead map the reference atlas incorrectly during registration, which undermines performance.

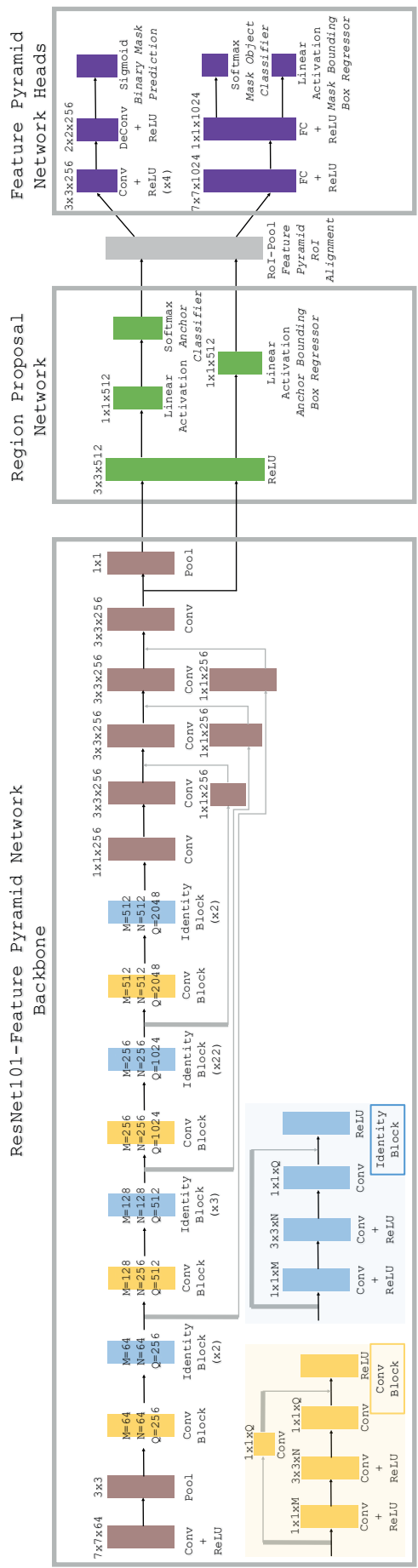

**Supplementary Fig. 8 | *SeBRe* neural network architecture.** A multistage Mask RCNN network is used, based on ResNet-101 and FPN as backbone architecture. The output of ResNet-FPN is fed to the RPN, where brain regions are localized and bounding boxes for RoIs proposed. The proposed feature maps are downsampled in the RoI pooling layer (RoIAlign layer), followed by the FPN heads where binary masks are proposed and brain regions are segmented pixelwise.

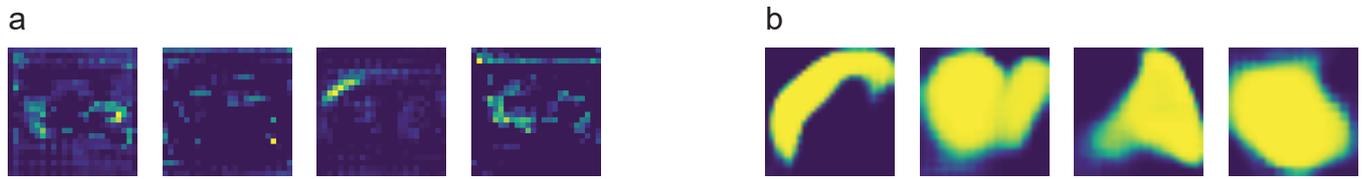

**Supplementary Fig. 9 | Output feature maps at different stages of *SeBRe* image processing pipeline. a**, Activations of feature maps produced by the ResNet101-FPN backbone architecture. **b**, Output feature maps of the FPN head, predicting the region-wise masks of brain sections.